\pdfoutput=1

\documentclass[11pt]{article}

\usepackage[]{acl}

\usepackage{times}
\usepackage{latexsym}
\usepackage{booktabs}
\usepackage{graphicx} 
\usepackage{algorithm} 
\usepackage[noend]{algpseudocode} 
\usepackage{multirow} 
\usepackage{arydshln}
\usepackage{amssymb}
\usepackage{pifont}
\newcommand{\cmark}{\ding{51}}%
\newcommand{\xmark}{\ding{55}}%
\newcommand\commentout[1]{}

\usepackage[T1]{fontenc}

\usepackage[utf8]{inputenc}

\usepackage{microtype}

\usepackage[normalem]{ulem}

\title{\emph{Turning Tables}: Generating Examples from Semi-structured Tables for Endowing Language Models with Reasoning Skills}

	\newif\ifcomments
\commentstrue 
\ifcomments
    \providecommand\at[1]{[\textcolor{blue}{{AT: #1}}]}
    \providecommand\jb[1]{[\textcolor{teal}{{JB: #1}}]}
    \providecommand\oy[1]{[\textcolor{violet}{{OY: #1}}]}
    \providecommand\es[1]{[\textcolor{orange}{{ES: #1}}]}
    \providecommand\ag[1]{[\textcolor{yellow}{{AG: #1}}]}
    
\else
    \providecommand\at[1]{}
    \providecommand\jb[1]{}
    \providecommand\oy[1]{}
    \providecommand\es[1]{}
    \providecommand\ag[1]{}

\fi

\newcommand{\uniformsampling}{uniform sampling}
\newcommand{\errorsampling}{error sampling}
\newcommand{\adaptivesampling}{momentum sampling}
\newcommand{\rem}{PReasM}
\newcommand{\remuniform}{\emph{PReasM-Uni}}
\newcommand{\remerrors}{\emph{PReasM-Err}}
\newcommand{\remadaptive}{\emph{PReasM-Moment}}
\newcommand{\tfive}{T5}
\newcommand{\fone}{F$_1$}

 \author{Ori Yoran$^{1}$ \Thanks{ Work done while working at the Allen Institute for Artificial Intelligence.} ~~ Alon Talmor$^{1}$\footnotemark[1] ~~ 
\textbf{Jonathan Berant}$^{1,2}$ \\ 
~$^1$Tel-Aviv University, $^2$The Allen Institute for AI \\
\texttt{\{oriy,alontalmor\}@mail.tau.ac.il}\\
\texttt{joberant@cs.tau.ac.il}\\
}

\begin{document}
\maketitle

\begin{abstract}
Models pre-trained with a language modeling objective possess ample world knowledge and language skills, but are known to struggle in tasks that require reasoning. 
In this work, we propose to leverage semi-structured tables, and automatically generate at scale question-paragraph pairs, where answering the question requires reasoning over multiple facts in the paragraph. We add a pre-training step over this synthetic data, which includes examples that require 16 different reasoning skills such as number comparison, conjunction, and fact composition. To improve data efficiency, we propose sampling strategies that focus training on reasoning skills the model is currently lacking. We evaluate our approach on three reading comprehension datasets that are focused on reasoning, and show that our model, \rem{}, substantially outperforms \tfive{}, a popular pre-trained encoder-decoder model. Moreover, sampling examples based on current model errors leads to faster training and higher overall performance.
\end{abstract}

\section{Introduction}

\begin{figure}[ht]
  \includegraphics[width=1.0\columnwidth]{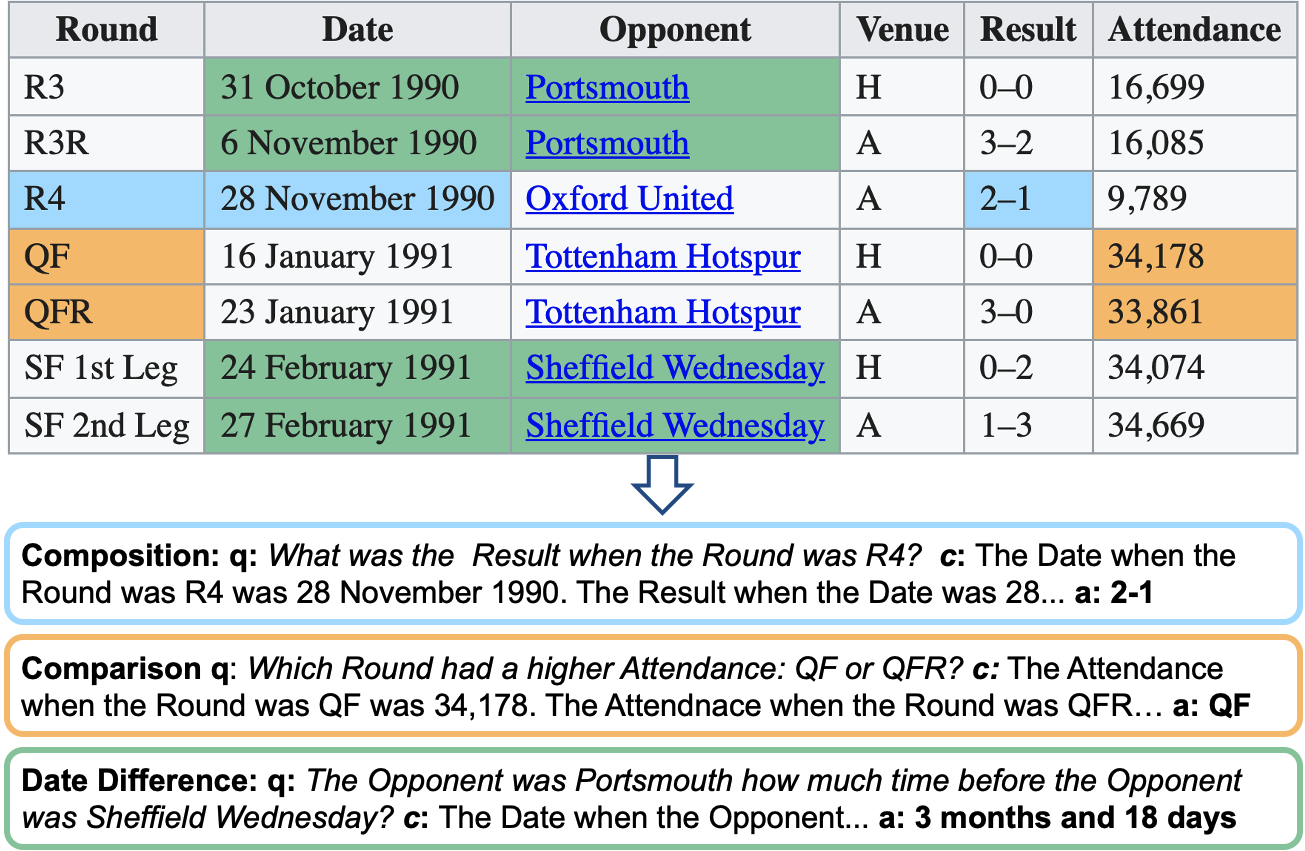}
  \caption{
  An example table and question-context-answer triplets generated from the table as synthetic data.  Each color corresponds to a different reasoning skill and colored cells are necessary to answer the question. The contexts shown are partial, such that the actual context contains the necessary information to answer the question and additional distractors. Answers are not necessarily extractive (e.g., date difference).
  }
  ~\label{fig:intro}
  
\end{figure}
Large pre-trained language models (LMs) \cite{devlin-etal-2019-bert, liu2019roberta, NEURIPS2020_1457c0d6,2020t5} have become the backbone of natural language processing in recent years. However, recent work has shown that they struggle in performing symbolic reasoning operations, such as composition or conjunction of facts \cite{talmor-etal-2019-commonsenseqa, talmor-etal-2020-olmpics}, numerical operations \cite{wallace-etal-2019-nlp, hidey-etal-2020-deseption}, and quantification \cite{ warstadt-etal-2019-investigating}, without substantial amounts of additional data.
 
Past work on improving reasoning skills in pre-trained models has taken two flavors: (a) adding specialized components for specific skills, like numerical and temporal reasoning \cite{ran-etal-2019-numnet, Gupta2020Neural, khot2021text, chen-etal-2020-question}, or (b) generating synthetic examples at scale, for example, by using grammars and templates  \cite{rozen-etal-2019-diversify, zhao-etal-2019-data, andreas-2020-good,asai-hajishirzi-2020-logic, campagna-etal-2020-zero, geva-etal-2020-injecting}, and question generation models \cite{alberti-etal-2019-synthetic, puri-etal-2020-training, bartolo2021improving}.



\begin{figure*}[hbt!]
  \includegraphics[width=2.0\columnwidth]{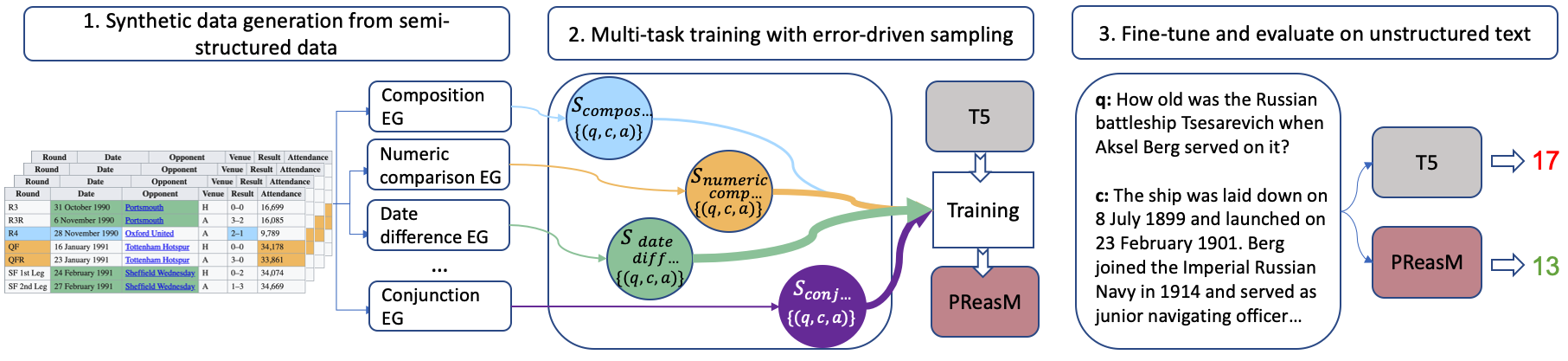}
  \caption{
    Approach overview.
    First, we use semi-structured tables to generate large amounts of data from 16 different example generators (EGs), each corresponding to a different reasoning skill. Then, a pre-trained LM is trained over this data in a multi-task setup to obtain our model, \rem{}, where we dynamically sample examples based on current model errors (arrow width corresponds to the number of sampled examples). Last, our model is fine-tuned and evaluated on target tasks that require reasoning. 
  }
  ~\label{fig:overview}
  
\end{figure*}

In this work, we take the latter approach 
and argue that semi-structured tables are a valuable resource for automatic generation of training data that will endow LMs with reasoning skills. Tables can be crawled from the web at scale, and cover a wide range of domains and topics. Moreover, their structured nature makes them amenable to automatic processes of data generation. Specifically, given a table, we use templates to generate reading comprehension (RC) examples, that is, question-context-answer triplets, where answering the question requires diverse types of reasoning over facts mentioned in the context. Fig.~\ref{fig:intro} shows an example table, and three generated question-context-answer examples, which require fact composition, number comparison, and computing a date difference. Unlike prior work where semi-structured data was used for reasoning over tables or knowledge-bases \cite{eisenschlos-etal-2020-understanding, yin-etal-2020-tabert,herzig-etal-2020-tapas, yu2021grappa}, here we harness tables to allow LMs to reason over \emph{text} directly.


Fig.~\ref{fig:overview} provides an overview of our approach.
We generate data by crawling tables from Wikipedia, and applying 16 different example generators (EGs) on each table.
Each EG corresponds to a particular reasoning skill (composition, numerical comparison, see Table~\ref{tab:templates-examples} for full list), and comprises a small set of question templates. Variables in the templates are filled with content from the table, and the structure of the table allows to compute the answer automatically. The context is a list of facts generated from the table that contain facts required for answering the question as well as distractor facts.


We add a pre-training step over this generated data, where we perform multi-task training over the 16 task corresponding to the EGs.
Since each EG can generate vast numbers of examples, 
it is important to focus training on reasoning skills that the model lacks. Thus, we use \emph{error-driven sampling} to construct training batches, where most examples are sampled from EGs that the model currently struggles with. 
We experiment with \emph{error sampling} \cite{gottumukkala-etal-2020-dynamic}, where examples are sampled in proportion to the current error rate on a particular task, and propose \emph{momentum sampling}, where examples are sampled in proportion to how fast the model is \emph{improving} on a task. We show that when some tasks are very noisy, momentum sampling is more data efficient than error sampling.


We fine tune our \textbf{P}re-traind for \textbf{Reas}oning \textbf{M}odel, \rem{}, on three RC datasets that require reasoning: DROP \cite{Dua2019DROP}, IIRC \cite{ferguson-etal-2020-iirc}, and MMQA \cite{talmor2021multimodalqa}. \rem{} outperforms the original pre-trained \tfive{} \cite{2020t5} model by significant margins: $7.6$, $4.1$, and $1.2$ \fone{} points, respectively. 
Our results set a new state-of-the-art on MMQA and are the best results on IIRC for models where the retriever and reader are trained separately. Our analysis shows that 
\rem{} leads to improvements of up to $40$ \fone{} points on specific question types, such as computing the difference between two dates, without causing a drop in other question types.

In conclusion, our results suggest that semi-structured tables are a viable and untapped source of information for automatically generating large amounts of data that can be used to endow LMs with reasoning skills that are not captured using current pre-training approaches.

Our code, data, and models are publicly available and can be downloaded from \url{https://github.com/oriyor/turning_tables}.


\section{Data Generation}\label{sec:data_generation}
We succinctly define the problem setup, and then turn to the process of automatic data generation from tables.

\paragraph{Problem Setup}

Our goal is to train a RC model that given a question $q$ and textual context $c$ returns an answer $a$ (Fig.~\ref{fig:intro}), given a training set $D = \{(q_i, c_i, a_i)\}_{i=1}^N$.
We focus on questions that require reasoning over the context $c$, e.g., composing two facts.
To endow LMs with reasoning skills, we would like to automatically generate a large synthetic training set $D_\textit{syn} = \{(q_j, c_j, a_j)\}_{j=1}^{M}$ (where $M \gg N$) from semi-structured tables, before fine-tuning on a target dataset.

\subsection{Generating Examples from Tables}
We use tables from English Wikipedia\footnote{We use the 01-01-2020 Wikipedia dump.} to generate $D_{\textit{syn}}$. English Wikipedia includes millions of tables 
with high lexical and domain diversity \cite{DBLP:journals/corr/abs-1902-01740, Chen2020TabFact, gupta-etal-2020-infotabs, talmor2021multimodalqa, DBLP:journals/corr/abs-2104-00369, DBLP:journals/corr/abs-2104-04243}.
We first extract from Wikipedia all  tables $\mathcal{T}$ that have at least two columns and 10-25 rows, resulting in more than 700K tables. Then, we annotate all table columns with their semantic type (\texttt{STRING}, \texttt{NUMBER}, or \texttt{DATE}), which allows us to generate questions that involve manipulating numbers and dates. We annotate the semantic type of each column with standard tools for parsing dates\footnote{\url{https://pypi.org/project/python-dateutil/}} and numbers.

\begin{table*}[t]
    \begin{center}
        \tiny
        \begin{tabular}{lll} \toprule
        {\emph{EG}} &  {\emph{Template}} & {\emph{Question}} \\ \midrule
        
        2/3-hop & What was the \texttt{col:1}(s) when the \texttt{col:2} was \texttt{val:2} in& ``What was the \textbf{Play(s)} when the \textbf{Author} was \textbf{William} \textbf{Shakespeare} in \textbf{Notable works}  \\
         Composition & \texttt{table-title} of \texttt{page-title}? & of \textbf{Lidia Zamkow?}'' \\
        \midrule
        
        Conjunction & What was the \texttt{col:1} when the \texttt{col:2} was \texttt{val:2} and the \texttt{col:3} & ``What was the \textbf{common name} when the \textbf{family} was \textbf{Picidae} and the \textbf{distribution} was  \\
        & was \texttt{val:3} in \texttt{table-title} of \texttt{page-title}?  & \textbf{Okinawa} in \textbf{List of species }of \textbf{List} \textbf{of endemic birds of Japan}?'' \\\midrule
        
        Quantifiers & Is \texttt{val:1} the only \texttt{col:1} that has \texttt{col:2} \texttt{val:2} in  \texttt{table-title}  & ``Is \textbf{Jean Philippe} the only \textbf{Artist} that has \textbf{Language French} in \textbf{Results} of \textbf{Eurovision} \\ 
        Only &  of \texttt{page-title}? &  \textbf{Song Contest 1959}?'' \\
        \midrule
        
        Quantifiers & In \texttt{table-title} of \texttt{page-title}, does [OPERATOR] \texttt{col:1}  & ``In \textbf{Coal} of \textbf{List of Mines in South Africa}, does \textbf{every Mine} have \textbf{Owner Exxaro}?'' \\ 
        Every/Most & have \texttt{col:2} \texttt{val:2}? &  \\
        \midrule
        
        Num. & In \texttt{col:1} of \texttt{table-title} of \texttt{page-title}, which \texttt{col:1} had & ``In \textbf{Administration} of \textbf{Mueang Nonthaburi District}, which \textbf{name} had \textbf{a higher} \\
        Comparison & [OPERATOR] \texttt{col:2}: \texttt{val:2} or \texttt{val:2}? &  \textbf{population: Suan Yai} or \textbf{Bang Khen?}'' \\ \midrule
        
        Temp. & In \texttt{col:1} of \texttt{table-title} of \texttt{page-title}, what happened  & ``In \textbf{Awards and nominations} of \textbf{Alexandre Pires}, what happened \textbf{earlier}: the \textbf{Category} \\
        Comparison & [OPERATOR]: the \texttt{col:1} was \texttt{val:1} or the \texttt{col:2} was \texttt{val:2}? & was \textbf{Pop New Artist} or the \textbf{Category} was \textbf{Album of the Year?}'' \\ \midrule
        
        Num. Boolean & In \texttt{col:1} of \texttt{table-title} of \texttt{page-title} did \texttt{val:1} have & ``In \textbf{Top employers} of \textbf{Chula Vista, California}, did \textbf{Walmart} have \textbf{more} \textbf{employees} \\
        Comparison & [OPERATOR] \texttt{col2} than \texttt{val:1}? & than \textbf{Target}?'' \\\midrule

        Temp. Boolean & The \texttt{col:1} was \texttt{val:1} [OPERATOR] the \texttt{col:2} was \texttt{val:2} in & 
        ``The \textbf{Referee} was \textbf{Salim Oussassi more recently than when} the \textbf{Referee} was \textbf{Rachid} \\
        Comparison & \texttt{table-title} of \texttt{page-title}? & \textbf{Medjiba} in \textbf{1980 to 1999} of \textbf{Algerian Cup Final referees?}'' \\\midrule

        Temp./Num. & In \texttt{table-title} of \texttt{page-title}, which \texttt{col:1} has the & ``In \textbf{List of graphic novels} of \textbf{Minx (comics)}, which \textbf{title} has the \textbf{earliest} \textbf{release date}?'' \\
        Superlatives & [OPERATOR] \texttt{col:2}? & \\\midrule

        Arithmetic & In \texttt{table-title} of \texttt{page-title}, what was the [OPERATOR] & ``In \textbf{By rocket} of \textbf{1961 in spaceflight}, what was the \textbf{highest Successes} when the \textbf{Remarks} \\
        Superlatives & \texttt{col:1} when the \texttt{col:2} was \texttt{val:2}? & was \textbf{Maiden flight?}'' \\\midrule
        
        Counting & How many \texttt{col:1} have \texttt{col:2} \texttt{val:2} in \texttt{table-title} of & ``How many \textbf{elections} have \textbf{candidate John Kufuor} in \textbf{Presidential elections} of \textbf{New} \\
        & \texttt{page-title}? & \textbf{Patriotic Party?}'' \\ \midrule
        
        Arithmetic & In \texttt{table-title} of \texttt{page-title}, what was the total number of & ``In \textbf{Assists table} of \textbf{2010-11 La Liga}, what was the total number of \textbf{assists} when the \\
        Addition & \texttt{col:1} when the \texttt{col:2} was \texttt{val2?}& \textbf{club} was \textbf{Villarreal?}'' \\\midrule
                
        Date & In \texttt{table-title} of \texttt{page-title}, how much time had passed bet-& ``In \textbf{Notable events | Concerts} of \textbf{Candlestick Park}, how much time had passed \\
        Difference & ween when the \texttt{col:1} was \texttt{val:1} and when the \texttt{col:2} was \texttt{val:2}? & between when the \textbf{Artist} was \textbf{Paul} \textbf{McCartney} and when the \textbf{Artist} was \textbf{The Beatles?}'' \\ \bottomrule
        
        \end{tabular} 
        \caption{Question templates with examples for all EGs. Variable names specify permissible instantiations, where \texttt{col} is a column name, \texttt{val} is a value, and indices denote that a value must originate from a particular column. 
        2/3-hop composition examples are generated by generating 2/3-long fact chains between the answer and the value in the question. For example, above, the chain will include the facts \emph{``The Role when the Author was Shakespeare was Lady Macbeth. The Play when the Role was Lady Macbeth was Macbeth''}.
        `[OPERATOR]' corresponds to EG-specific operators that we instantiate, e.g., in the EG `Temp. comparison' [OPERATOR] is replaced with \emph{`earlier'} or \emph{`later'}. Some EGs are collapsed into a single row (e.g., Quantifiers Every/Most). 
        }
        \label{tab:templates-examples}
    \end{center}
\end{table*}

The core of the generation process are the example generators (EGs), each corresponding to a particular reasoning skill, such as composition, conjunction, etc. (see Table~\ref{tab:templates-examples}). 
Each example generator $g \in \mathcal{G}$ is a function that takes a table $t \in \mathcal{T}$
and randomly samples ten $(q,c,a)$ triplets from the set of all possible triplets, where (i) $q$ is a question is pseudo-language, (ii) $c$ is the context, i.e., a list of facts extracted from $t$ that includes the \emph{gold facts} necessary for answering $q$ and \emph{distractor facts}, all phrased in pseudo-language, and (iii) $a$ is the answer. 
Overall, the synthetic training set is $D_{\textit{syn}}=\bigcup_{t \in \mathcal{T}} \bigcup_{g \in \mathcal{G}} g(t)$.


EGs generate examples in the following way. Each EG is associated with one or more question templates, which differ in their surface phrasing. Templates contain typed variables that are instantiated with content from the table (see all variables in Table~\ref{tab:templates-examples}). Column and value variables are indexed to specify that the variable \texttt{val:i} must be instantiated by a value from the column \texttt{col:i}. In temporal/numerical templates some column and value variables must have a \texttt{DATE}/\texttt{NUMBER} type, but we omit this notation for brevity. Instantiating all variables results in the question $q$ and the template allows us to programmatically compute the answer $a$. 
E.g., in the question from Fig.~\ref{fig:intro}: \emph{``In League Cup of 1990–91 Chelsea F.C. season, Which Round had a higher Attendance: QF or QFR?''} the answer $a$ can be found by finding the rows with the values \emph{``QF''} and \emph{``QFR''} in the column \emph{``Round''}, and returning the value that has a higher number in the column \emph{``Attendance''}. 

The context $c$ is generated from the table content necessary for answering the question, which can be identified using the instantiated question template.
Facts generally have the form ``The \texttt{col:1} when the \texttt{col:2} was \texttt{val:2} was \texttt{val:1}''.
For example, to answer the question above, we generate the gold facts \emph{``The Attendance when the Round was QF was 34,178''}
\emph{``The Attendance when the Round was QFR was 33,861''} using the relevant column names and values.
We also generate additional distractor facts by generating facts from rows or columns that are not relevant for the question, for example, \emph{``The Attendance when the Round was R4 was 9,789''}. Finally, we shuffle the gold facts and distractor facts. 

Overall, our process results in a large set $D_\textit{syn}$, which includes examples that require reasoning from 16 EGs (all shown in Table~\ref{tab:templates-examples}).

\subsection{Data Analysis}

The data generation process yields $4.8M$ questions from over $176K$ tables, and their main statistics are in Table~\ref{tab:key-statistics}. 
The number of distinct words and word pieces is very large (850K and 27K respectively), illustrating the wide coverage and high lexical diversity of our approach. Moreover, generated examples have diverse answer types, which include extracting spans from the context, yes/no questions, numeric, and date answers. In addition, by leveraging the distribution of Wikipedia tables our questions cover a wide range of domains including popular culture, geography, politics and science. 
Specifically, tables cover more than 2,500 different Wikipedia categories, with 150 categories covering 80\% of the data. We show the most frequent categories in \S\ref{sec:sup-dg}.

\section{Training}\label{sec:training}



Since our EGs generate large quantities of examples, one can think of each EG as providing an infinite stream of examples.
In this setup, a natural question is how to construct training batches such that the model learns the required skills as quickly as possible. After briefly describing our model, we will detail our training framework, where we sample examples from EGs in an error-driven manner.

\paragraph{Model} We use a standard encoder-decoder architecture \cite{2020t5, lewis-etal-2020-bart}. Given a training example $(q,c,a)$, the model takes as input the sequence of tokens `$q \textrm{ [SEP] } c$', and the task is to autoregressively decode the answer $a$ token-by-token.
We train to maximize the maximum likelihood objective $\log P(a\mid q,c)$.

\subsection{Multi-task Training over Reasoning Skills}

Given a pre-trained LM, we add another pre-training step, where we multi-task over a set of tasks $\mathcal{S}$, where each task corresponds to examples generated from a single EG.
Similar to past work \cite{yogatama2019learning, geva-etal-2020-injecting}, to avoid ``catastrophic forgetting'' \cite{DBLP:journals/corr/KirkpatrickPRVD16} of the language skills acquired during pre-training, we sample batches from the original pre-training task with probability $\lambda=0.5$.

Past work \cite{gottumukkala-etal-2020-dynamic} has shown that \emph{heterogeneous batching}, i.e., having examples from all tasks in each batch, leads to better performance compared to having entire batches from a single task. We follow this practice, and in every batch
sample examples from every task according to a probability distribution $P_\textit{tasks} \in \mathbb{R}^{|\mathcal{S}|}$.
The main question is how to determine the distribution $P_\textit{tasks}$, which we turn to next.

\begin{table}[t]
\centering
        \footnotesize
        \begin{tabular}{ll} \toprule
        {\emph{Measurement}} & {\emph{Value}}  \\ \midrule
        \# Distinct Questions & 4.8M \\
        \# Distinct tables & 176K \\
        \# Distinct pages & 130K \\ \midrule
        Avg. question length (words) & 19.3$\pm$4.2  \\  
        Avg. context length (words) & 111.3$\pm$44.8  \\ \midrule
        Avg. \# of gold facts & 4.4$\pm$4.7  \\  
        Avg. \# of distractors facts & 5.0$\pm$2.8  \\ \midrule
        \# Distinct words & 850,543 \\
        \# Distinct word-pieces & 27,055 \\ \midrule
        \% Span answer & 43.2 \\
        \% Yes/no answer & 31.6 \\
        \% Numeric answer & 15.8 \\
        \% Date answer & 9.4 \\  \bottomrule
        \end{tabular} 
\caption{Key statistics for the generated data. 
}
\label{tab:key-statistics}
\end{table}


\subsection{Sampling Strategies}

We describe strategies for computing $P_\textit{tasks}$, starting with the commonly-used \emph{uniform sampling} approach, and then turn to error-driven approaches.

\paragraph{Uniform sampling} Past work \cite{khashabi-etal-2020-unifiedqa, 2020t5, wang-etal-2020-balancing} used uniform sampling, where the probability to sample from a task $s$ is $P_\textit{tasks}(s) = \frac{1}{|\mathcal{S}|}$, as a-priori all tasks are equally important.
Other approaches sample examples in proportion to the size of the training set \cite{2020t5, wang-etal-2020-balancing}. This is not applicable in our case, where we assume an infinite stream of examples for every task, and also make no assumptions on the distribution over reasoning skills in the downstream test set.

\paragraph{Error sampling}
Recent work \cite{sharma2017learning, gottumukkala-etal-2020-dynamic} 
has proposed to construct $P_\textit{tasks}$ based on model errors, where one over-samples tasks where the error rate is high. 
More formally, let $\textit{Ceil}(s)$ be an estimate of the maximum accuracy achievable on a task $s$, and $\textit{Acc}(s)$ be the current model accuracy for task $s$ on an held-out set. We define $\Delta(s)=\textit{Ceil}(s)-\textit{Acc}(s)$ and  $P_\textit{tasks}(s) = \frac{\Delta(s)}{\sum_{s'}\Delta(s')}$. The distribution $P_\textit{tasks}$ is updated every time we evaluate the current model on the held-out data.
In our setup, since we perform multi-task training over synthetic examples that are generated at scale, we assume that $\forall_{s \in \mathcal{S}}\textit{Ceil}(s)=1.0$ and hence: $\Delta(s)=\textit{Err}(s) = 1.0-\textit{Acc}(s)$.

\paragraph{Momentum Sampling}

An issue with \errorsampling{} is that if the error rate is high for a task and learning it is slow, the model will spend most time on that task at the expense of all other tasks, which may lead overall to low data efficiency. We empirically demonstrate this phenomenon at the bottom of this section. To remedy this phenomenon, we introduce a new sampling strategy, termed \textit{\adaptivesampling{}}.

In \adaptivesampling{}, we sample examples from a task in proportion to its \emph{rate of improvement}, putting most probability mass on skills that are improving quickly.
Alg.~\ref{alg:adaptive} provides the details of this strategy. Let $t$ denote the index of a checkpoint evaluated on the held-out set, let $w$ be a window size, and $\textit{Acc}_s(i)$ be the held-out accuracy of checkpoint $i$ on task $s$. We estimate model accuracy on a task $s$ at the beginning and end of the window, and sample examples in proportion to the difference\footnote{We use the difference in performance and not the gain to account for cases of sudden drops in performance.} in accuracy during that window. To smooth out accuracy fluctuations in adjacent checkpoints, we estimate accuracy as an average of $k$ model checkpoints. During the first $w$ checkpoint evaluations, we simply use uniform sampling.

A desired property of \adaptivesampling{} is that when all tasks reach their ceiling accuracy, it converges to uniform sampling, unlike error sampling that will over-sample from tasks for which $\textit{Ceil}(s)$ is low. This is beneficial in cases where there is variance in $\textit{Ceil}(s)$ across tasks. We illustrate this point empirically next.

\begin{algorithm}[t]
    \small
	\caption{Momentum Sampling($w$, $\epsilon, k, t$)} 
	\textbf{Input:} windows size $w$, smoothing factor $k$, minimum share of examples per task $\epsilon$, training time $t$.
	\begin{algorithmic}[1]
         
	    \For {$s \in \mathcal{S}$}
            
            \If{$t \geq w$}
            
    		    \State $\textit{Acc}_{\textrm{head}}$ $ \leftarrow \frac{1}{k}\sum_{i=t-k}^{t} \textit{Acc}_s(i)$ 
    		    \State $\textit{Acc}_{\textrm{tail}}$ $\leftarrow{}$ $\frac{1}{k}\sum_{i=t-w}^{t-w+k} \textit{Acc}_s(i) $
    		    \State $P_\textit{tasks}[s]$ $\leftarrow{}$ $\max(|\textit{Acc}_{\textrm{head}}-\textit{Acc}_\textrm{tail}|, \epsilon)$ 
		    
		    \Else{}
		        
		        \State $P_\textit{tasks}[s]$ $\leftarrow{}$ $1/|\mathcal{S}|$
		   
		   \EndIf

		\EndFor
	\State $P_\textit{tasks} \leftarrow{} P_\textit{tasks}/\| P_\textit{tasks} \|_{1}$ 
	\end{algorithmic} 
	\label{alg:adaptive}
\end{algorithm}

\paragraph{Empirical comparison of sampling strategies}

To highlight the benefits of \adaptivesampling{}, we show that when sampling from two tasks, where the labels for one of the tasks are noisy, \adaptivesampling{} outperforms \errorsampling{}.
Specifically, we consider training on 2-hop composition and arithmetic addition, which is slower to train, in two conditions: (a) with gold labels for both tasks, and (b) when the labels for the 2-hop composition are randomly sampled from the vocabulary. We expect that when labels for 2-hop composition are random, this will lead to slow training of arithmetic addition when using \errorsampling{}, since most of the probability mass will be dedicated to 2-hop composition, which is impossible to learn. 

\begin{figure}
  \includegraphics[width=1.0\columnwidth]{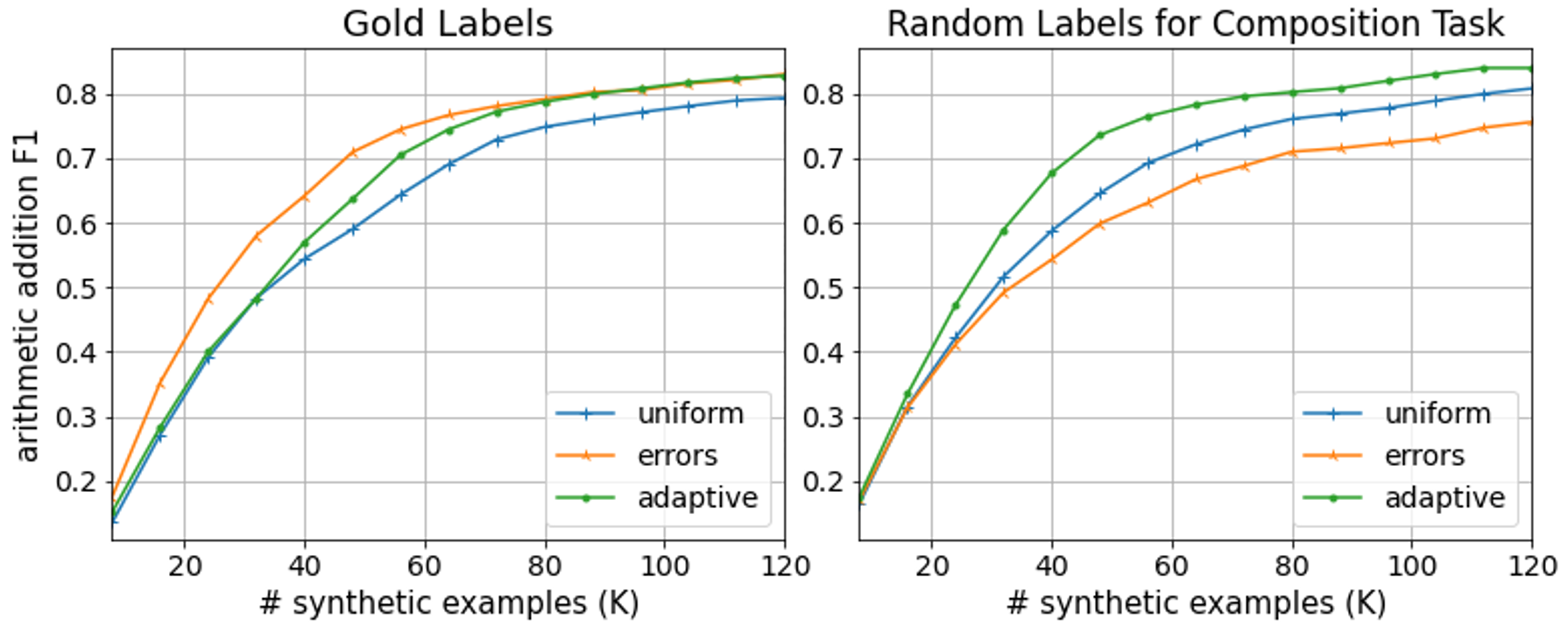}
  \caption{Motivation for momentum sampling. With the gold labels (left), \errorsampling{} and \adaptivesampling{} outperform \uniformsampling{} on the arithmetic addition task by over-sampling the harder task. When 2-hop composition has random labels (right), \errorsampling{} over-samples the composition task and momentum sampling is best.
  }
  ~\label{fig:sampling-motivation}
\end{figure}

Fig.~\ref{fig:sampling-motivation} illustrates this phenomenon. Without noise (left), both \adaptivesampling{} and \errorsampling{} learn faster than uniform sampling. Momentum sampling learns more slowly than \errorsampling{}, due to the warm-start period in the first $w$ evaluated checkpoints. However, when 2-hop composition has random labels (right), \errorsampling{} puts most probability mass on 2-hop composition, and thus \errorsampling{} is even worse than uniform sampling, while \adaptivesampling{} performs best. Thus, \adaptivesampling{} outperforms uniform sampling in both cases.

\paragraph{Related work}


Past work has considered error-driven data sampling in the context of active learning \cite{sharma2017learning}, reinforcement learning \cite{graves2017automated, glover2019task, xu-etal-2019-multi}, transfer learning \cite{zhang2020worstcaseaware, pilault2021conditionally}, and distributionally robust optimization \cite{oren-etal-2019-distributionally, Sagawa*2020Distributionally}, where the goal is to perform well over a family of distributions over the tasks. Similar to \newcite{gottumukkala-etal-2020-dynamic}, we compute $P_\textit{tasks}$ based on accuracy over a held-out set rather than the loss over a training data, as this corresponds directly to our target metric.

\section{Experimental Setup}\label{sec:experimental_setup}


We now describe our experimental evaluation.

\subsection{Models}

\paragraph{Baselines}

Our baseline is \tfive{} \cite{2020t5}, a popular pre-trained encoder-decoder model, which we fine-tune on the downstream datasets. 
We experiment with two model sizes, 220 million parameters (\tfive{-Base}), and 770 million parameters (\tfive{-Large}). When the answer is a list, we train our model to generate the list of values. 

Our pre-trained for reasoning model, \rem{}, is a \tfive{} model where we add a second step of pre-training on $D_\textit{syn}$. Again, we experiment with Base and Large models and three sampling strategies: \uniformsampling{}, \errorsampling{}, and \adaptivesampling; we name our models  \remuniform{}, \remerrors{}, and \remadaptive{} accordingly. 

\subsection{Datasets}

\paragraph{DROP}
\cite{Dua2019DROP} is a RC dataset with questions that require mathematical reasoning. As an additional baseline, we also compare to GenBERT \cite{geva-etal-2020-injecting}, which similar to our approach injects numerical skills by automatically generating synthetic data from a grammar.


\paragraph{IIRC} \cite{ferguson-etal-2020-iirc} is a question answering dataset, where annotators were given a single Wikipedia paragraph, and were asked to author questions that depend on that paragraph, but also on other paragraphs linked from the input paragraph, without observing the said paragraphs.
This resulted in questions that require discrete temporal ($28\%$) or numeric ($11\%$) reasoning. In addition, $30\%$ of the questions are unanswerable. 

We experiment with IIRC in both an \emph{oracle} and \emph{retrieval} setting. In the oracle setting, the model is given the gold context, which reduces the problem to reading comprehension, where we can apply our models.
In the retrieval setting, we use the improved pipeline model introduced by \newcite{DBLP:journals/corr/abs-2103-12235} to retrieve the relevant context, and then replace the NumNet+ (Base) reader \cite{ran-etal-2019-numnet} used by the authors (which has specialized architecture for numerical reasoning) with \tfive{}/\rem{}.

\paragraph{MMQA} 

\cite{talmor2021multimodalqa} 
is a question answering dataset, where the input is a question and a context that consists of a table, multiple text paragraphs, and multiple images, and the model must reason over a subset of the input modalities to answer the question.\footnote{We removed tables that appear in the MMQA development and test sets from $D_{\textit{syn}}$.} Since our \tfive{}/\rem{} models cannot handle images or very long contexts, we construct a pipeline that automatically directs some MMQA questions to \tfive{}/\rem{}, and uses the original \emph{Implicit-Decomp} baseline from \newcite{talmor2021multimodalqa} elsewhere.

The first classifier in this pipeline is a \tfive{-large} model fine-tuned on the MMQA training set to determine if a question is likely to require an image or not. When the classifier determines a question requires an image, the example is directed to \emph{Implicit-Decomp}. The accuracy of this classifier on the MMQA development set is $99.2$\%.

The second classifier in the pipeline is a \tfive{-3B} model, fine-tuned on the MMQA training set to determine given a question and one of the textual paragraphs if that paragraph is required for answering the question.
Then, for every question that does not require an image, we classify each of the textual paragraphs and only use the ones classified as relevant. This process identifies all gold paragraphs in $95.8$\% of the examples. Again, we experiment with an \emph{oracle} and \emph{retrieval} setting, such that in the oracle setting our model is presented with the gold text paragraphs.

Last, we convert the table into text by linearizing the table as described in \newcite{talmor2021multimodalqa}. The model is presented with multiple paragraphs and the linearized table, and can answer questions that require any reasoning across them. Since the context is long, we  present the model with contexts of size 1,536 word-pieces (without any change to the original \tfive{} model).

Table~\ref{tab:datasets} contains the number of questions in the train, development, and test sets for each of our datasets. For MMQA, there are 15,688 train and 1,501 development examples that require reasoning over the table and text only.

\begin{table}\resizebox{1.0\columnwidth}{!}{
\centering
        \tiny
        \begin{tabular}{lccc} \toprule
        {\emph{Dataset}} & {\emph{\# Train}} & {\emph{\# Development}} & {\emph{\# Test}} \\
        {\emph{}} & {\emph{Questions}} & {\emph{Questions}} & {\emph{Questions}}  \\ 
        \midrule
        DROP & 77,409 & 9,536 & 9,622 \\
        IIRC & 10,839 & 1,301 & 1,301 \\ 
        MMQA & 23,817 & 2,441 & 3,660 \\\bottomrule{}
    \end{tabular}}
\caption{Number of questions in each dataset. 
}
\label{tab:datasets}
\end{table}

\paragraph{Evaluation metrics}
For all datasets, we use the F$_1$ and EM scores defined in DROP \cite{Dua2019DROP}, and later used in IIRC and MMQA, where
given a gold and predicted list of answers, items on the two lists are aligned, and then their strings are compared. We use the official evaluation scripts released by the dataset authors in all cases.

\section{Experimental Results}

We first present results on the downstream RC datasets (\S\ref{subsec:downstream_res}) and then examine performance directly on the synthetic data (\S\ref{subsec:synthetic_res}).


\subsection{Performance on RC Datasets}
\label{subsec:downstream_res}


Table~\ref{tab:test-results} presents the results of our large models over all datasets, also in comparison to current state-of-the-art. We observe that \rem{} substantially improves performance compared to \tfive{} in all conditions, improving on the test set by $7.6$, $7.9$, $4.1$, and $1.2$ \fone{} points on DROP, IIRC$_\textrm{oracle}$, IIRC , and MMQA respectively. We obtain new state-of-the-art results on MMQA and IIRC$_\textrm{oracle}$. On IIRC, we improve performance when using the same retriever (Pipeline) and replacing the NumNet+ retriever with \rem{}.\footnote{We report the official numbers from \newcite{DBLP:journals/corr/abs-2103-12235} ($45.8$/$44.3$ F$_1$ on the development/test sets). To fairly compare with the NumNet+ reader, we got the retrieved paragraphs for the Pipeline model through personal communication. However, results on these paragraphs was lower than reported in the paper: $45.5$/$42.8$ F$_1$. The reported results of our models are with this slightly worse retriever, which still outperforms the performance of NumNet+ (Pipeline) as reported in the original paper.} On DROP, specialized architectures for handling numbers still substantially outperform both \tfive{} and \rem{}. 

Table~\ref{tab:rem-comparison} shows the effect of different sampling strategies when training the \rem{} model. We observe that error sampling and momentum sampling generally outperform uniform sampling, but do not observe a clear advantage to momentum sampling compared to error sampling. We further analyze the effects of momentum sampling and error sampling when pre-training on $D_\textit{syn}$ in \S\ref{subsec:synthetic_res}.

We now look in detail at the performance of models on different answer types across datasets, where we observe that \rem{} leads to dramatic improvements on some types, while maintaining similar performance on other types.


\begin{table}\resizebox{1.0\columnwidth}{!}{

\centering

        \tiny
        \begin{tabular}{llcc} \toprule
        
        \emph{Dataset} & \emph{Model} & \emph{Development} & \emph{Test}   \\ \midrule
        
        \multirow{4}{*}{DROP} & \tfive{-Large} & 64.6/61.8 & 65.0/61.8 \\
         & \rem{-Large} & \textbf{72.3/69.4} &  \textbf{72.6/69.5} \\ \cdashline{2-4}
         & GenBERT & 72.3/68.8 &  72.4/68.6 \\ 
         & QDGAT-ALBERT &  &  \textbf{90.1/87.0} \\

         \midrule
         
        \multirow{3}{*}{IIRC$_\textrm{oracle}$} & \tfive{-Large} & 69.9/64.9 & 67.1/62.7 \\
         & \rem{-Large} &  \textbf{77.4/72.7}& \textbf{75.0/70.6} \\ \cdashline{2-4}
         & NumNet+ & 69.2/63.9 &  70.3/65.6 \\ 

         \midrule
         
        \multirow{4}{*}{IIRC} & \tfive{-Large} (Pipeline) & 47.4/44.2 & 41.0/37.8 \\
         & \rem{-Large} (Pipeline) & \textbf{50.0/46.5} & \textbf{45.1/42.0} \\ \cdashline{2-4}
         & NumNet+ (Pipeline) & 45.8/41.7 &  44.3/41.3 \\ 
         & NumNet+ (Joint) & \textbf{50.6/46.9} &  \textbf{50.5/47.4} \\

         \midrule
        
        \multirow{3}{*}{MMQA} & \tfive{-Large} & 64.3/57.9 & 63.4/57.0 \\ 
         & \rem{-Large} & \textbf{65.5/59.0} & \textbf{64.6/58.3} \\ \cdashline{2-4}
         & \emph{Implicit-Decomp} & 55.5/48.8 &  55.9/49.3 \\ 

        \bottomrule
        \end{tabular}}
\caption{Development and test results. The two values in each cell indicate \fone{}/EM. 
} 
\label{tab:test-results}
\end{table}


\begin{table} \resizebox{1.0\columnwidth}{!}{
\centering
        \tiny
        \begin{tabular}{lcccc} \toprule

        \emph{Model} &  \emph{DROP} & \emph{IIRC$_\textrm{oracle}$} & \emph{IIRC} &  \emph{MMQA} \\\midrule
        
        \tfive{-Base} & 55.4$\pm0.3$ & 65.8$\pm0.3$ &  43.4$\pm0.1$ & 61.4$\pm0.2$  \\ 
        \remuniform{-Base} & 67.4$\pm0.2$  & 72.9$\pm0.3$ & \textbf{47.3$\pm0.3$} & \textbf{62.7$\pm0.2$}  \\ 
         \remadaptive{-Base} & 67.7$\pm0.1$ & 73.4$\pm0.3$ & 46.8$\pm0.2$ & 62.5$\pm0.1$ \\ 
         \remerrors{-Base} & \textbf{68.0$\pm0.1$}  & \textbf{74.3$\pm0.2$} & 47.0$\pm0.3$ & 62.6 $\pm0.3$  \\ 
        \midrule

         \tfive{-Large} &  64.6$\pm0.1$ &  69.7$\pm0.2$ & 47.0$\pm0.3$ & 64.2$\pm0.2$  \\     
         \remuniform{-Large} & 71.4$\pm0.1$ &  75.0$\pm0.2$ & 49.1$\pm0.2$ & 64.9$\pm0.4$ \\ 
         \remadaptive{-Large} & 71.7$\pm0.1$ &  \textbf{76.8$\pm0.5$} &\textbf{49.8$\pm0.2$} & 64.9$\pm0.2$ \\ 
         \remerrors{-Large}  & \textbf{72.2$\pm0.1$} & 76.1$\pm0.3$ & 49.2$\pm0.5$ & \textbf{65.3$\pm0.1$} \\

        \bottomrule
        \end{tabular}}
\caption{F$_1$ on the development set with different sampling strategies. Results show the average and standard deviation over 3 seeds. 
}
\label{tab:rem-comparison}
\end{table}

\paragraph{DROP} 
\rem{} outperforms \tfive{} by $12.6$ points in the Base model and by $7.6$ points in the Large model (see Table~\ref{tab:rem-comparison}). Table~\ref{tab:drop-breakdown} breaks down performance based on answer types, where we see that \rem{} outperforms \tfive{} across the board for all model sizes and answer types by a large margin.


Comparing \rem{}-Base to GenBERT, which is a Base size model, we find \rem{} outperforms GenBERT, on 3 of the 4 answer types.
The high performance of GenBERT on \emph{Number} questions can be explained by several factors: (a) GenBERT uses digit tokenization which improves arithmetic reasoning \cite{thawani-etal-2021-representing}, and (b) training on multiple templates that focus on numerical reasoning. 
Training \rem{} on more numeric data generated from the grammar of GenBERT is likely to lead to further improvements.


\begin{table}[t]\resizebox{1.0\columnwidth}{!}{
\centering
        \tiny
        \begin{tabular}{lccccc} \toprule
        
        \emph{Model} & \emph{Span} & \emph{Spans} & \emph{Date} & \emph{Number} & \emph{Total }  \\\midrule
         \tfive{-Base} & 77.5 & 65.8 & 57.1 & 43.7 & 55.8 \\
         \rem{-Base} & \textbf{81.1} & \textbf{69.4} & \textbf{64.6} & \textbf{61.5} & \textbf{68.1} \\ \midrule
         \tfive{-Large}  & 86.1 & \textbf{78.4} & 75.7 & 52.2 & 64.6 \\
         \rem{-Large} & \textbf{86.6} & \textbf{78.4} & \textbf{77.7}
         & \textbf{64.4} &  \textbf{72.3} \\ \midrule
         GenBERT & 74.5 & 24.2 & 56.4 & \textbf{75.2} &  \textbf{72.3} \\
        \bottomrule
        \end{tabular}}
        
\caption{Development F$_1$ on DROP with answer type breakdown.}
\label{tab:drop-breakdown}
\end{table}

\paragraph{IIRC} 
Table~\ref{tab:iirc-breakdown} breaks down performance based on answer types. Again, \rem{} outperforms \tfive{} in the oracle setup by roughly $8$ points for both Base and Large models, and by $2.6$-$4$ points in the retrieval setup.
Improvements are mostly due to cases when the answer is a numerical \emph{Value}, where \rem{} outperforms \tfive{} by $39.1$ and $40.3$ \fone{} points in Base and Large models (oracle setup). 

Comparing \rem{}-Base to NumNet+, we find that \rem{} outperforms NumNet+ on \emph{None}, \emph{Span} and \emph{Binary} questions, but has lower performance on \emph{Value} questions, where NumNet+ uses specialized architecture.

Uniform sampling slightly outperforms error-driven sampling in the Base model on IIRC (Table~\ref{tab:rem-comparison}). Analyzing answer types, we find that error-driven sampling improves performance on \emph{Value} questions, but reduces performance on \emph{None} questions, leading overall to a slight advantage for uniform sampling. This effect disappears in Large models, where error-driven sampling outperforms uniform sampling.



Overall, \rem{}-Large improves state-of-the-art in the oracle setup by $4.7$ F$_1$ points. In the retrieval setting, \rem{} outperforms NumNet+ (Pipeline) by $4.2$ and $0.8$ \fone{} points on the development and test sets, respectively. 

\commentout{
\rem{-Large} achieves a new state-of-the-art in the oracle and .
In the oracle setting, we outperform the NumNet+ results reported in \newcite{ferguson-etal-2020-iirc} by 4.7\fone{} points. In the pipeline retrieval setting, we outperform the improved NumNet+ model introduced in \newcite{DBLP:journals/corr/abs-2103-12235} by $4.2$ and $0.8$ \fone{} points in the development and test sets. 
}

\begin{table}\resizebox{1.0\columnwidth}{!}{
\centering
        \tiny
        \begin{tabular}{lcccccc} \toprule
        
        \emph{Model} &  \emph{Oracle} & \emph{None} & \emph{Span} & \emph{Binary}  & \emph{Value}  & \emph{Total} \\\midrule
         \tfive{-Base} & \cmark & 91.4 & 72.0 & 76.6 & 8.7  & 66.3 \\
         \rem{-Base} & \cmark & 92.5 & 74.9 & 71.9 & 47.8 & \textbf{74.5}\\ \midrule
         \tfive{-Large} & \cmark  & 92.2 & 77.7 & 81.3 & 10.9 & 69.9 \\
         \rem{-Large} & \cmark & 92.2 & 78.4 & 80.5 & 51.2 & \textbf{77.4} \\ \midrule
         \tfive{-Base} & \xmark & 57.1 & 47.6 & 54.7 & 6.7 & 43.5 \\
         \rem{-Base} & \xmark & 53.9 & 49.1 & 64.8 & 24.3 & \textbf{47.5} \\ \midrule
         \tfive{-Large} & \xmark & 56.2 & 49.9 & 77.3 & 11.5  & 47.4 \\
         \rem{-Large} & \xmark & 55.9 & 50.8 & 69.5 & 28.6 & \textbf{50.0} \\ \midrule
         NumNet+ (Pipeline) & \xmark & 49.6 & 48.4 & 52.3 & 30.0 & 45.8 \\ 
        \bottomrule 
         
        \end{tabular}}
\caption{Development F$_1$ on IIRC with answer type breakdown. } 
\label{tab:iirc-breakdown}
\end{table}

\paragraph{MMQA}

\begin{table*}
\centering
        \tiny
        \begin{tabular}{llcccccccc} \toprule
        \emph{Model} & \emph{Oracle} & \emph{ColumnHop} & \emph{Text} & \emph{Composition} & \emph{Comparison} & \emph{Conjunction} & \emph{Yes/No} & \emph{Aggregate} & \emph{Total} \\ \midrule
        \tfive{-Base} & \xmark & 81.7 & 75.2 & 67.0 & 61.8 & 74.1 & 76.9 & 27.3 & 71.9 \\
        \rem{-Base} & \xmark & 80.8 & 75.7 & 66.3 & \textbf{80.8} & \textbf{80.8} & \textbf{83.1} & \textbf{36.4} & \textbf{74.3} \\ \midrule
        \tfive{-Large} & \xmark & 82.6 & 79.8 & 71.8 & 69.3 & 83.0 & 83.1 & 27.3 & 76.8 \\
        \rem{-Large} & \xmark & 84.0 & 79.7 & 71.9 & \textbf{81.0} & 82.3 & \textbf{93.8} & \textbf{36.4} & \textbf{78.4} \\ \midrule
        \tfive{-Base} & \cmark & 85.2 & 82.1 & 74.6 & 63.3 & 77.4 & 80.0 & 27.3 & 77.9 \\
        \rem{-Base} & \cmark & 86.9 & 80.0 & 75.4 & \textbf{84.1} & \textbf{82.6} & \textbf{89.2} & \textbf{36.4} & \textbf{79.9} \\ \midrule
        \tfive{-Large} & \cmark & 88.2 & 85.9 & 79.4 & 74.1 & 83.2 & 83.1 & 36.4 & 82.7 \\
        \rem{-Large} & \cmark & 87.8 & 85.6 & 79.8 & \textbf{83.6} & 82.3 & \textbf{90.8} & \textbf{45.5} & \textbf{83.8} \\ \midrule
        \emph{Implicit-Decomp} & \cmark & \textbf{96.6} & 57.1 & 53.2 & 78.4 & 68.1 & 76.9 & \textbf{59.1} & 62.3 \\ \midrule
        \end{tabular} 
\caption{Development \fone{} on MMQA with reasoning type breakdown on the development set.
}
\label{tab:mmqa-results} 
\end{table*}

Table~\ref{tab:mmqa-results} breaks down model performance based on reasoning skill, which is annotated for every example in MMQA.
\rem{} outperforms \tfive{} in both the oracle and retrieval setting, and for both model sizes. 

We observe that the main cause of improvement are comparison questions, where \rem{} outperforms \tfive{} by $19$ and $11.7$ \fone{} on Base and Large models. Second, \rem{} outperforms \tfive{} on questions that require conjunction in Base models, and yes/no questions in all settings. Interestingly, \tfive{} is equipped with decent composition skills, \emph{without} any specialized pre-training and based only on the original T5 pre-training.

Comparing our models to \emph{Implicit-Decomp}, we find that although \emph{Implicit-Decomp} outperforms our models on questions that require hopping between two table columns and performing aggregations (there are only 11 aggregation questions in the development set), \rem{} outperforms \emph{Implicit-Decomp} in all other  cases. When considering only questions that require reasoning over text and tables, \rem{}-Large improves \fone{} by $16.1$ points, from $62.3$ to $78.4$.


        
         

\subsection{Perofrmance on $D_\textit{syn}$}
\label{subsec:synthetic_res}

\begin{figure*}[ht!]
  \centering
  \includegraphics[width=2.0\columnwidth]{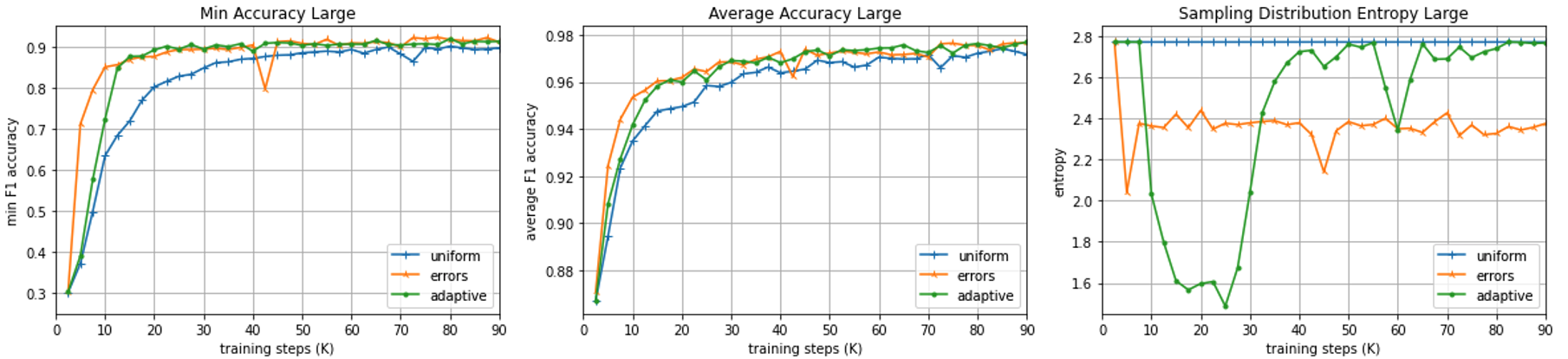}
  \caption{
   Minimum and average task accuracy over a held-out set from $D_\textit{syn}$ (left and center), and the entropy of $P_\textit{tasks}$ (right), for Large models, as a function of the number of training steps for all sampling strategies.
  }
  ~\label{fig:syn-results}
\end{figure*}

Fig.~\ref{fig:syn-results} shows statistics on the performance of \rem{} on different tasks in $D_\textit{syn}$ during training. 
The average accuracy across all 16 tasks at the end of training is high -- almost 98.0 accuracy.
We observe that \rem{} reaches high performance on all tasks, where the lowest-performing tasks are `arithmetic addition' and `date difference', where the accuracy is at most $91.8$ and $95.0$ respectively at the end of training. On those tasks, the advantage of error-driven sampling is evident, and it outperforms uniform sampling by as much as 4 points. We provide full results over $D_\textit{syn}$, including the performance of \tfive{} in a few-shot setting in \S\ref{sec:sup-results}.

Zooming-in on the learning curve, we see that momentum and error sampling learn reasoning skills a lot faster than uniform sampling.
Looking at the entropy of $P_\textit{tasks}$ sheds light on the difference between error sampling and momentum sampling. Error sampling tends to put most probability mass on the lowest-performing task, namely arithmetic addition, and thus its entropy over tasks is roughly constant from a certain point in training. Conversely, momentum sampling puts a lot of probability mass on tasks that are improving quickly at the beginning of training, but as improvements plateau, it converges towards uniform sampling.


\subsection{Analyzing Reasoning Skills in DROP}

\begin{table}\resizebox{1.0\columnwidth}{!}{
\centering
        \tiny
        \begin{tabular}{lccccc} \toprule
        
        \emph{Question Type}  & \emph{NMN} & \tfive{-} & \rem{-} & \tfive{-} & \rem{-} \\
        & & \emph{Base} & \emph{Base} & \emph{Large} & \emph{Large}  \\ \midrule 
        Date-Compare & 82.6 & 86.4 & 87.5 & 87.6 & \textbf{89.9} \\
        Date-Difference & 75.4 & 19.6 & 78.9 & 45.4 & \textbf{80.4} \\
        Number-Compare & 92.7 & 91.3 & 95.2 & 97.3 & \textbf{98.5} \\
        Extract-Number & 86.1 & 91.8 & 94.9 & 92.1 & \textbf{95.1} \\
        Count & 55.7 & 80.1 & 86.7 & 86.7 & \textbf{89.2} \\
        Extract-Argument & 69.7 & 87.6 & 86.2 & 90.5 & \textbf{92.1} \\
        \bottomrule 
         
        \end{tabular}}
\caption{F$_1$ on a previously-proposed split of a subset of the development set of DROP to reasoning skills.} 
\label{tab:analysis-endowed-skills}
\end{table}

To check which reasoning skills \rem{} has, we use a proposed split of a subset of DROP to reasoning skills \cite{Gupta2020Neural}. Table~\ref{tab:analysis-endowed-skills} presents the \fone{} results for our best \rem{} and \tfive{} models on this split, as well as the \fone{} results from the neural module network (NMN) used in \newcite{Gupta2020Neural}. We note that NMN were trained only on a subset of the original DROP dataset.
When comparing to \tfive{}, we find that \rem{} dramatically improves performance Date-Difference, and also leads to sizable gains in Number-Compare, Extract-Number and Count. In addition, \rem{} outperforms NMN on all reasoning skills.

\section{Related Work}
\label{sec:related}

\paragraph{Data augmenation}
Data augmentation techniques have been extensively explored in reading comprehension, question answering, and dialogue \cite{feng2021survey}, mainly by transfer learning \cite{talmor-berant-2019-multiqa, khashabi-etal-2020-unifiedqa} and synthetic data generation \cite{wei2018fast, zhao-etal-2019-data, alberti-etal-2019-synthetic,rozen-etal-2019-diversify,campagna-etal-2020-zero,Chen2020Neural, asai-hajishirzi-2020-logic, andreas-2020-good, puri-etal-2020-training,Asai2020Learning,geva-etal-2020-injecting, yang2021nt5,  bartolo2021improving}. Here we focus on semi-structured data as a valuable resource for data generation.

\paragraph{Pre-training over semi-structured data}
Past work on pre-training over tables focused on reasoning over tables and knowledge-bases \cite{eisenschlos-etal-2020-understanding, yin-etal-2020-tabert,herzig-etal-2020-tapas, 2021-tapas, yu2021grappa, neeraja-etal-2021-incorporating}, while we focus on reasoning over text.
Recently, \newcite{DBLP:journals/corr/abs-2106-01074} introduced a new dataset that focuses on reasoning over synthetic textual facts, which are generated by a LM from a knowledge graph.

  
\section{Conclusion}
In this work, we propose semi-structured tables as a valuable resource for automatically generating at scale examples that can endow pre-trained language models with reasoning skills.
We generate almost 5M examples that correspond to 16 reasoning skills from Wikipedia tables and add a second pre-training step over this data. To improve data efficiency we use error-driven sampling, which focuses training on reasoning skills that the model currently lacks.

We evaluate our model, \rem{}, on three reasoning-focused RC datasets and show that it leads to substantial improvements in all cases. Moreover, we thoroughly analyze the performance of \rem{} and show that our approach dramatically improves performance on questions that require reasoning skills that were not acquired during the original pre-training, while maintaining comparable performance on other question types.   

\section*{Acknowledgments}
We thank Elad Segal and Ankit Gupta for their useful comments and James Ferguson, Ansong Ni and Matt Gardner for their help with the IIRC dataset. This research was partially supported by The Yandex Initiative for Machine Learning, and the European Research Council (ERC) under the European
Union Horizons 2020 research and innovation programme (grant ERC DELPHI 802800).

\bibliography{anthology,custom}

\begin{thebibliography}{61}
\expandafter\ifx\csname natexlab\endcsname\relax\def\natexlab#1{#1}\fi

\bibitem[{Alberti et~al.(2019)Alberti, Andor, Pitler, Devlin, and
  Collins}]{alberti-etal-2019-synthetic}
Chris Alberti, Daniel Andor, Emily Pitler, Jacob Devlin, and Michael Collins.
  2019.
\newblock \href {https://doi.org/10.18653/v1/P19-1620} {Synthetic {QA} corpora
  generation with roundtrip consistency}.
\newblock In \emph{Proceedings of the 57th Annual Meeting of the Association
  for Computational Linguistics}, pages 6168--6173, Florence, Italy.
  Association for Computational Linguistics.

\bibitem[{Andreas(2020)}]{andreas-2020-good}
Jacob Andreas. 2020.
\newblock \href {https://doi.org/10.18653/v1/2020.acl-main.676} {Good-enough
  compositional data augmentation}.
\newblock In \emph{Proceedings of the 58th Annual Meeting of the Association
  for Computational Linguistics}, pages 7556--7566, Online. Association for
  Computational Linguistics.

\bibitem[{Asai and Hajishirzi(2020)}]{asai-hajishirzi-2020-logic}
Akari Asai and Hannaneh Hajishirzi. 2020.
\newblock \href {https://doi.org/10.18653/v1/2020.acl-main.499} {Logic-guided
  data augmentation and regularization for consistent question answering}.
\newblock In \emph{Proceedings of the 58th Annual Meeting of the Association
  for Computational Linguistics}, pages 5642--5650, Online. Association for
  Computational Linguistics.

\bibitem[{Asai et~al.(2020)Asai, Hashimoto, Hajishirzi, Socher, and
  Xiong}]{Asai2020Learning}
Akari Asai, Kazuma Hashimoto, Hannaneh Hajishirzi, Richard Socher, and Caiming
  Xiong. 2020.
\newblock \href {https://openreview.net/forum?id=SJgVHkrYDH} {Learning to
  retrieve reasoning paths over wikipedia graph for question answering}.
\newblock In \emph{8th International Conference on Learning Representations,
  {ICLR} 2020, Addis Ababa, Ethiopia, April 26-30, 2020}. OpenReview.net.

\bibitem[{Bartolo et~al.(2021)Bartolo, Thrush, Jia, Riedel, Stenetorp, and
  Kiela}]{bartolo2021improving}
Max Bartolo, Tristan Thrush, Robin Jia, Sebastian Riedel, Pontus Stenetorp, and
  Douwe Kiela. 2021.
\newblock \href {http://arxiv.org/abs/2104.08678} {Improving question answering
  model robustness with synthetic adversarial data generation}.

\bibitem[{Brown et~al.(2020)Brown, Mann, Ryder, Subbiah, Kaplan, Dhariwal,
  Neelakantan, Shyam, Sastry, Askell, Agarwal, Herbert{-}Voss, Krueger,
  Henighan, Child, Ramesh, Ziegler, Wu, Winter, Hesse, Chen, Sigler, Litwin,
  Gray, Chess, Clark, Berner, McCandlish, Radford, Sutskever, and
  Amodei}]{NEURIPS2020_1457c0d6}
Tom~B. Brown, Benjamin Mann, Nick Ryder, Melanie Subbiah, Jared Kaplan,
  Prafulla Dhariwal, Arvind Neelakantan, Pranav Shyam, Girish Sastry, Amanda
  Askell, Sandhini Agarwal, Ariel Herbert{-}Voss, Gretchen Krueger, Tom
  Henighan, Rewon Child, Aditya Ramesh, Daniel~M. Ziegler, Jeffrey Wu, Clemens
  Winter, Christopher Hesse, Mark Chen, Eric Sigler, Mateusz Litwin, Scott
  Gray, Benjamin Chess, Jack Clark, Christopher Berner, Sam McCandlish, Alec
  Radford, Ilya Sutskever, and Dario Amodei. 2020.
\newblock \href
  {https://proceedings.neurips.cc/paper/2020/hash/1457c0d6bfcb4967418bfb8ac142f64a-Abstract.html}
  {Language models are few-shot learners}.
\newblock In \emph{Advances in Neural Information Processing Systems 33: Annual
  Conference on Neural Information Processing Systems 2020, NeurIPS 2020,
  December 6-12, 2020, virtual}.

\bibitem[{Campagna et~al.(2020)Campagna, Foryciarz, Moradshahi, and
  Lam}]{campagna-etal-2020-zero}
Giovanni Campagna, Agata Foryciarz, Mehrad Moradshahi, and Monica Lam. 2020.
\newblock \href {https://doi.org/10.18653/v1/2020.acl-main.12} {Zero-shot
  transfer learning with synthesized data for multi-domain dialogue state
  tracking}.
\newblock In \emph{Proceedings of the 58th Annual Meeting of the Association
  for Computational Linguistics}, pages 122--132, Online. Association for
  Computational Linguistics.

\bibitem[{Chen et~al.(2020{\natexlab{a}})Chen, Xu, Cheng, Xiaochuan, Zhang,
  Song, Wang, Qi, and Chu}]{chen-etal-2020-question}
Kunlong Chen, Weidi Xu, Xingyi Cheng, Zou Xiaochuan, Yuyu Zhang, Le~Song,
  Taifeng Wang, Yuan Qi, and Wei Chu. 2020{\natexlab{a}}.
\newblock \href {https://doi.org/10.18653/v1/2020.emnlp-main.549} {Question
  directed graph attention network for numerical reasoning over text}.
\newblock In \emph{Proceedings of the 2020 Conference on Empirical Methods in
  Natural Language Processing (EMNLP)}, pages 6759--6768, Online. Association
  for Computational Linguistics.

\bibitem[{Chen et~al.(2020{\natexlab{b}})Chen, Wang, Chen, Zhang, Wang, Li,
  Zhou, and Wang}]{Chen2020TabFact}
Wenhu Chen, Hongmin Wang, Jianshu Chen, Yunkai Zhang, Hong Wang, Shiyang Li,
  Xiyou Zhou, and William~Yang Wang. 2020{\natexlab{b}}.
\newblock \href {https://openreview.net/forum?id=rkeJRhNYDH} {Tabfact: {A}
  large-scale dataset for table-based fact verification}.
\newblock In \emph{8th International Conference on Learning Representations,
  {ICLR} 2020, Addis Ababa, Ethiopia, April 26-30, 2020}. OpenReview.net.

\bibitem[{Chen et~al.(2020{\natexlab{c}})Chen, Liang, Yu, Zhou, Song, and
  Le}]{Chen2020Neural}
Xinyun Chen, Chen Liang, Adams~Wei Yu, Denny Zhou, Dawn Song, and Quoc~V. Le.
  2020{\natexlab{c}}.
\newblock \href {https://openreview.net/forum?id=ryxjnREFwH} {Neural symbolic
  reader: Scalable integration of distributed and symbolic representations for
  reading comprehension}.
\newblock In \emph{8th International Conference on Learning Representations,
  {ICLR} 2020, Addis Ababa, Ethiopia, April 26-30, 2020}. OpenReview.net.

\bibitem[{Devlin et~al.(2019)Devlin, Chang, Lee, and
  Toutanova}]{devlin-etal-2019-bert}
Jacob Devlin, Ming-Wei Chang, Kenton Lee, and Kristina Toutanova. 2019.
\newblock \href {https://doi.org/10.18653/v1/N19-1423} {{BERT}: Pre-training of
  deep bidirectional transformers for language understanding}.
\newblock In \emph{Proceedings of the 2019 Conference of the North {A}merican
  Chapter of the Association for Computational Linguistics: Human Language
  Technologies, Volume 1 (Long and Short Papers)}, pages 4171--4186,
  Minneapolis, Minnesota. Association for Computational Linguistics.

\bibitem[{Dua et~al.(2019)Dua, Wang, Dasigi, Stanovsky, Singh, and
  Gardner}]{Dua2019DROP}
Dheeru Dua, Yizhong Wang, Pradeep Dasigi, Gabriel Stanovsky, Sameer Singh, and
  Matt Gardner. 2019.
\newblock \href {https://doi.org/10.18653/v1/N19-1246} {{DROP}: A reading
  comprehension benchmark requiring discrete reasoning over paragraphs}.
\newblock In \emph{Proceedings of the 2019 Conference of the North {A}merican
  Chapter of the Association for Computational Linguistics: Human Language
  Technologies, Volume 1 (Long and Short Papers)}, pages 2368--2378,
  Minneapolis, Minnesota. Association for Computational Linguistics.

\bibitem[{Eisenschlos et~al.(2020)Eisenschlos, Krichene, and
  M{\"u}ller}]{eisenschlos-etal-2020-understanding}
Julian Eisenschlos, Syrine Krichene, and Thomas M{\"u}ller. 2020.
\newblock \href {https://doi.org/10.18653/v1/2020.findings-emnlp.27}
  {Understanding tables with intermediate pre-training}.
\newblock In \emph{Findings of the Association for Computational Linguistics:
  EMNLP 2020}, pages 281--296, Online. Association for Computational
  Linguistics.

\bibitem[{Feng et~al.(2021)Feng, Gangal, Wei, Chandar, Vosoughi, Mitamura, and
  Hovy}]{feng2021survey}
Steven~Y. Feng, Varun Gangal, Jason Wei, Sarath Chandar, Soroush Vosoughi,
  Teruko Mitamura, and Eduard Hovy. 2021.
\newblock \href {http://arxiv.org/abs/2105.03075} {A survey of data
  augmentation approaches for {NLP}}.

\bibitem[{Ferguson et~al.(2020)Ferguson, Gardner, Hajishirzi, Khot, and
  Dasigi}]{ferguson-etal-2020-iirc}
James Ferguson, Matt Gardner, Hannaneh Hajishirzi, Tushar Khot, and Pradeep
  Dasigi. 2020.
\newblock \href {https://doi.org/10.18653/v1/2020.emnlp-main.86} {{IIRC}: A
  dataset of incomplete information reading comprehension questions}.
\newblock In \emph{Proceedings of the 2020 Conference on Empirical Methods in
  Natural Language Processing (EMNLP)}, pages 1137--1147, Online. Association
  for Computational Linguistics.

\bibitem[{Fetahu et~al.(2019)Fetahu, Anand, and
  Koutraki}]{DBLP:journals/corr/abs-1902-01740}
Besnik Fetahu, Avishek Anand, and Maria Koutraki. 2019.
\newblock \href {https://doi.org/10.1145/3308558.3313629} {Tablenet: An
  approach for determining fine-grained relations for wikipedia tables}.
\newblock In \emph{The World Wide Web Conference, {WWW} 2019, San Francisco,
  CA, USA, May 13-17, 2019}, pages 2736--2742. {ACM}.

\bibitem[{Geva et~al.(2020)Geva, Gupta, and Berant}]{geva-etal-2020-injecting}
Mor Geva, Ankit Gupta, and Jonathan Berant. 2020.
\newblock \href {https://doi.org/10.18653/v1/2020.acl-main.89} {Injecting
  numerical reasoning skills into language models}.
\newblock In \emph{Proceedings of the 58th Annual Meeting of the Association
  for Computational Linguistics}, pages 946--958, Online. Association for
  Computational Linguistics.

\bibitem[{Glover and Hokamp(2019)}]{glover2019task}
John Glover and Chris Hokamp. 2019.
\newblock \href {http://arxiv.org/abs/1907.06214} {Task selection policies for
  multitask learning}.

\bibitem[{Gottumukkala et~al.(2020)Gottumukkala, Dua, Singh, and
  Gardner}]{gottumukkala-etal-2020-dynamic}
Ananth Gottumukkala, Dheeru Dua, Sameer Singh, and Matt Gardner. 2020.
\newblock \href {https://doi.org/10.18653/v1/2020.acl-main.86} {Dynamic
  sampling strategies for multi-task reading comprehension}.
\newblock In \emph{Proceedings of the 58th Annual Meeting of the Association
  for Computational Linguistics}, pages 920--924, Online. Association for
  Computational Linguistics.

\bibitem[{Graves et~al.(2017)Graves, Bellemare, Menick, Munos, and
  Kavukcuoglu}]{graves2017automated}
Alex Graves, Marc~G. Bellemare, Jacob Menick, R{\'{e}}mi Munos, and Koray
  Kavukcuoglu. 2017.
\newblock \href {http://proceedings.mlr.press/v70/graves17a.html} {Automated
  curriculum learning for neural networks}.
\newblock In \emph{Proceedings of the 34th International Conference on Machine
  Learning, {ICML} 2017, Sydney, NSW, Australia, 6-11 August 2017}, volume~70
  of \emph{Proceedings of Machine Learning Research}, pages 1311--1320. {PMLR}.

\bibitem[{Gupta et~al.(2020{\natexlab{a}})Gupta, Lin, Roth, Singh, and
  Gardner}]{Gupta2020Neural}
Nitish Gupta, Kevin Lin, Dan Roth, Sameer Singh, and Matt Gardner.
  2020{\natexlab{a}}.
\newblock \href {https://openreview.net/forum?id=SygWvAVFPr} {Neural module
  networks for reasoning over text}.
\newblock In \emph{8th International Conference on Learning Representations,
  {ICLR} 2020, Addis Ababa, Ethiopia, April 26-30, 2020}. OpenReview.net.

\bibitem[{Gupta et~al.(2020{\natexlab{b}})Gupta, Mehta, Nokhiz, and
  Srikumar}]{gupta-etal-2020-infotabs}
Vivek Gupta, Maitrey Mehta, Pegah Nokhiz, and Vivek Srikumar.
  2020{\natexlab{b}}.
\newblock \href {https://doi.org/10.18653/v1/2020.acl-main.210} {{INFOTABS}:
  Inference on tables as semi-structured data}.
\newblock In \emph{Proceedings of the 58th Annual Meeting of the Association
  for Computational Linguistics}, pages 2309--2324, Online. Association for
  Computational Linguistics.

\bibitem[{Herzig et~al.(2020)Herzig, Nowak, M{\"u}ller, Piccinno, and
  Eisenschlos}]{herzig-etal-2020-tapas}
Jonathan Herzig, Pawel~Krzysztof Nowak, Thomas M{\"u}ller, Francesco Piccinno,
  and Julian Eisenschlos. 2020.
\newblock \href {https://doi.org/10.18653/v1/2020.acl-main.398} {{T}a{P}as:
  Weakly supervised table parsing via pre-training}.
\newblock In \emph{Proceedings of the 58th Annual Meeting of the Association
  for Computational Linguistics}, pages 4320--4333, Online. Association for
  Computational Linguistics.

\bibitem[{Hidey et~al.(2020)Hidey, Chakrabarty, Alhindi, Varia, Krstovski,
  Diab, and Muresan}]{hidey-etal-2020-deseption}
Christopher Hidey, Tuhin Chakrabarty, Tariq Alhindi, Siddharth Varia, Kriste
  Krstovski, Mona Diab, and Smaranda Muresan. 2020.
\newblock \href {https://doi.org/10.18653/v1/2020.acl-main.761}
  {{D}e{S}e{P}tion: Dual sequence prediction and adversarial examples for
  improved fact-checking}.
\newblock In \emph{Proceedings of the 58th Annual Meeting of the Association
  for Computational Linguistics}, pages 8593--8606, Online. Association for
  Computational Linguistics.

\bibitem[{Khashabi et~al.(2020)Khashabi, Min, Khot, Sabharwal, Tafjord, Clark,
  and Hajishirzi}]{khashabi-etal-2020-unifiedqa}
Daniel Khashabi, Sewon Min, Tushar Khot, Ashish Sabharwal, Oyvind Tafjord,
  Peter Clark, and Hannaneh Hajishirzi. 2020.
\newblock \href {https://doi.org/10.18653/v1/2020.findings-emnlp.171}
  {{UNIFIEDQA}: Crossing format boundaries with a single {QA} system}.
\newblock In \emph{Findings of the Association for Computational Linguistics:
  EMNLP 2020}, pages 1896--1907, Online. Association for Computational
  Linguistics.

\bibitem[{Khot et~al.(2021)Khot, Khashabi, Richardson, Clark, and
  Sabharwal}]{khot2021text}
Tushar Khot, Daniel Khashabi, Kyle Richardson, Peter Clark, and Ashish
  Sabharwal. 2021.
\newblock \href {https://www.aclweb.org/anthology/2021.naacl-main.99} {Text
  modular networks: Learning to decompose tasks in the language of existing
  models}.
\newblock In \emph{Proceedings of the 2021 Conference of the North American
  Chapter of the Association for Computational Linguistics: Human Language
  Technologies}, pages 1264--1279, Online. Association for Computational
  Linguistics.

\bibitem[{Kirkpatrick et~al.(2016)Kirkpatrick, Pascanu, Rabinowitz, Veness,
  Desjardins, Rusu, Milan, Quan, Ramalho, Grabska{-}Barwinska, Hassabis,
  Clopath, Kumaran, and Hadsell}]{DBLP:journals/corr/KirkpatrickPRVD16}
James Kirkpatrick, Razvan Pascanu, Neil~C. Rabinowitz, Joel Veness, Guillaume
  Desjardins, Andrei~A. Rusu, Kieran Milan, John Quan, Tiago Ramalho, Agnieszka
  Grabska{-}Barwinska, Demis Hassabis, Claudia Clopath, Dharshan Kumaran, and
  Raia Hadsell. 2016.
\newblock \href {http://arxiv.org/abs/1612.00796} {Overcoming catastrophic
  forgetting in neural networks}.
\newblock \emph{CoRR}, abs/1612.00796.

\bibitem[{Lewis et~al.(2020)Lewis, Liu, Goyal, Ghazvininejad, Mohamed, Levy,
  Stoyanov, and Zettlemoyer}]{lewis-etal-2020-bart}
Mike Lewis, Yinhan Liu, Naman Goyal, Marjan Ghazvininejad, Abdelrahman Mohamed,
  Omer Levy, Veselin Stoyanov, and Luke Zettlemoyer. 2020.
\newblock \href {https://doi.org/10.18653/v1/2020.acl-main.703} {{BART}:
  Denoising sequence-to-sequence pre-training for natural language generation,
  translation, and comprehension}.
\newblock In \emph{Proceedings of the 58th Annual Meeting of the Association
  for Computational Linguistics}, pages 7871--7880, Online. Association for
  Computational Linguistics.

\bibitem[{Liu et~al.(2019)Liu, Ott, Goyal, Du, Joshi, Chen, Levy, Lewis,
  Zettlemoyer, and Stoyanov}]{liu2019roberta}
Yinhan Liu, Myle Ott, Naman Goyal, Jingfei Du, Mandar Joshi, Danqi Chen, Omer
  Levy, Mike Lewis, Luke Zettlemoyer, and Veselin Stoyanov. 2019.
\newblock \href {http://arxiv.org/abs/1907.11692} {Roberta: A robustly
  optimized bert pretraining approach}.

\bibitem[{Müller et~al.(2021)Müller, Eisenschlos, and Krichene}]{2021-tapas}
Thomas Müller, Julian~Martin Eisenschlos, and Syrine Krichene. 2021.
\newblock \href {http://arxiv.org/abs/2104.01099} {Tapas at semeval-2021 task
  9: Reasoning over tables with intermediate pre-training}.

\bibitem[{Nan et~al.(2021)Nan, Hsieh, Mao, Lin, Verma, Zhang, Kryscinski,
  Schoelkopf, Kong, Tang, Mutuma, Rosand, Trindade, Bandaru, Cunningham, Xiong,
  and Radev}]{DBLP:journals/corr/abs-2104-00369}
Linyong Nan, Chiachun Hsieh, Ziming Mao, Xi~Victoria Lin, Neha Verma, Rui
  Zhang, Wojciech Kryscinski, Nick Schoelkopf, Riley Kong, Xiangru Tang, Murori
  Mutuma, Ben Rosand, Isabel Trindade, Renusree Bandaru, Jacob Cunningham,
  Caiming Xiong, and Dragomir~R. Radev. 2021.
\newblock \href {http://arxiv.org/abs/2104.00369} {Fetaqa: Free-form table
  question answering}.
\newblock \emph{CoRR}, abs/2104.00369.

\bibitem[{Neeraja et~al.(2021{\natexlab{a}})Neeraja, Gupta, and
  Srikumar}]{DBLP:journals/corr/abs-2104-04243}
J.~Neeraja, Vivek Gupta, and Vivek Srikumar. 2021{\natexlab{a}}.
\newblock \href {https://www.aclweb.org/anthology/2021.naacl-main.224}
  {Incorporating external knowledge to enhance tabular reasoning}.
\newblock In \emph{Proceedings of the 2021 Conference of the North American
  Chapter of the Association for Computational Linguistics: Human Language
  Technologies}, pages 2799--2809, Online. Association for Computational
  Linguistics.

\bibitem[{Neeraja et~al.(2021{\natexlab{b}})Neeraja, Gupta, and
  Srikumar}]{neeraja-etal-2021-incorporating}
J.~Neeraja, Vivek Gupta, and Vivek Srikumar. 2021{\natexlab{b}}.
\newblock \href {https://www.aclweb.org/anthology/2021.naacl-main.224}
  {Incorporating external knowledge to enhance tabular reasoning}.
\newblock In \emph{Proceedings of the 2021 Conference of the North American
  Chapter of the Association for Computational Linguistics: Human Language
  Technologies}, pages 2799--2809, Online. Association for Computational
  Linguistics.

\bibitem[{Ni et~al.(2021)Ni, Gardner, and
  Dasigi}]{DBLP:journals/corr/abs-2103-12235}
Ansong Ni, Matt Gardner, and Pradeep Dasigi. 2021.
\newblock \href {http://arxiv.org/abs/2103.12235} {Mitigating false-negative
  contexts in multi-document questionanswering with retrieval marginalization}.
\newblock \emph{CoRR}, abs/2103.12235.

\bibitem[{Oren et~al.(2019)Oren, Sagawa, Hashimoto, and
  Liang}]{oren-etal-2019-distributionally}
Yonatan Oren, Shiori Sagawa, Tatsunori Hashimoto, and Percy Liang. 2019.
\newblock \href {https://doi.org/10.18653/v1/D19-1432} {Distributionally robust
  language modeling}.
\newblock In \emph{Proceedings of the 2019 Conference on Empirical Methods in
  Natural Language Processing and the 9th International Joint Conference on
  Natural Language Processing (EMNLP-IJCNLP)}, pages 4227--4237, Hong Kong,
  China. Association for Computational Linguistics.

\bibitem[{Pilault et~al.(2021)Pilault, hattami, and
  Pal}]{pilault2021conditionally}
Jonathan Pilault, Amine~El hattami, and Christopher Pal. 2021.
\newblock \href {https://openreview.net/forum?id=de11dbHzAMF} {Conditionally
  adaptive multi-task learning: Improving transfer learning in {NLP} using
  fewer parameters {\&} less data}.
\newblock In \emph{International Conference on Learning Representations}.

\bibitem[{Puri et~al.(2020)Puri, Spring, Shoeybi, Patwary, and
  Catanzaro}]{puri-etal-2020-training}
Raul Puri, Ryan Spring, Mohammad Shoeybi, Mostofa Patwary, and Bryan Catanzaro.
  2020.
\newblock \href {https://doi.org/10.18653/v1/2020.emnlp-main.468} {Training
  question answering models from synthetic data}.
\newblock In \emph{Proceedings of the 2020 Conference on Empirical Methods in
  Natural Language Processing (EMNLP)}, pages 5811--5826, Online. Association
  for Computational Linguistics.

\bibitem[{Raffel et~al.(2020)Raffel, Shazeer, Roberts, Lee, Narang, Matena,
  Zhou, Li, and Liu}]{2020t5}
Colin Raffel, Noam Shazeer, Adam Roberts, Katherine Lee, Sharan Narang, Michael
  Matena, Yanqi Zhou, Wei Li, and Peter~J. Liu. 2020.
\newblock \href {http://jmlr.org/papers/v21/20-074.html} {Exploring the limits
  of transfer learning with a unified text-to-text transformer}.
\newblock \emph{Journal of Machine Learning Research}, 21(140):1--67.

\bibitem[{Ram et~al.(2021)Ram, Kirstain, Berant, Globerson, and
  Levy}]{ram2021fewshotqa}
Ori Ram, Yuval Kirstain, Jonathan Berant, Amir Globerson, and Omer Levy. 2021.
\newblock Few-shot question answering by pretraining span selection.
\newblock In \emph{Association for Computational Linguistics (ACL)}.

\bibitem[{Ran et~al.(2019)Ran, Lin, Li, Zhou, and Liu}]{ran-etal-2019-numnet}
Qiu Ran, Yankai Lin, Peng Li, Jie Zhou, and Zhiyuan Liu. 2019.
\newblock \href {https://doi.org/10.18653/v1/D19-1251} {{N}um{N}et: Machine
  reading comprehension with numerical reasoning}.
\newblock In \emph{Proceedings of the 2019 Conference on Empirical Methods in
  Natural Language Processing and the 9th International Joint Conference on
  Natural Language Processing (EMNLP-IJCNLP)}, pages 2474--2484, Hong Kong,
  China. Association for Computational Linguistics.

\bibitem[{Rozen et~al.(2019)Rozen, Shwartz, Aharoni, and
  Dagan}]{rozen-etal-2019-diversify}
Ohad Rozen, Vered Shwartz, Roee Aharoni, and Ido Dagan. 2019.
\newblock \href {https://doi.org/10.18653/v1/K19-1019} {Diversify your
  datasets: Analyzing generalization via controlled variance in adversarial
  datasets}.
\newblock In \emph{Proceedings of the 23rd Conference on Computational Natural
  Language Learning (CoNLL)}, pages 196--205, Hong Kong, China. Association for
  Computational Linguistics.

\bibitem[{Sagawa et~al.(2020)Sagawa, Koh, Hashimoto, and
  Liang}]{Sagawa*2020Distributionally}
Shiori Sagawa, Pang~Wei Koh, Tatsunori~B. Hashimoto, and Percy Liang. 2020.
\newblock \href {https://openreview.net/forum?id=ryxGuJrFvS} {Distributionally
  robust neural networks}.
\newblock In \emph{8th International Conference on Learning Representations,
  {ICLR} 2020, Addis Ababa, Ethiopia, April 26-30, 2020}. OpenReview.net.

\bibitem[{Sharma et~al.(2018)Sharma, Jha, Hegde, and
  Ravindran}]{sharma2017learning}
Sahil Sharma, Ashutosh~Kumar Jha, Parikshit Hegde, and Balaraman Ravindran.
  2018.
\newblock \href {https://openreview.net/forum?id=B1nZ1weCZ} {Learning to
  multi-task by active sampling}.
\newblock In \emph{6th International Conference on Learning Representations,
  {ICLR} 2018, Vancouver, BC, Canada, April 30 - May 3, 2018, Conference Track
  Proceedings}. OpenReview.net.

\bibitem[{Talmor and Berant(2019)}]{talmor-berant-2019-multiqa}
Alon Talmor and Jonathan Berant. 2019.
\newblock \href {https://doi.org/10.18653/v1/P19-1485} {{M}ulti{QA}: An
  empirical investigation of generalization and transfer in reading
  comprehension}.
\newblock In \emph{Proceedings of the 57th Annual Meeting of the Association
  for Computational Linguistics}, pages 4911--4921, Florence, Italy.
  Association for Computational Linguistics.

\bibitem[{Talmor et~al.(2020)Talmor, Elazar, Goldberg, and
  Berant}]{talmor-etal-2020-olmpics}
Alon Talmor, Yanai Elazar, Yoav Goldberg, and Jonathan Berant. 2020.
\newblock \href {https://doi.org/10.1162/tacl_a_00342} {o{LM}pics-on what
  language model pre-training captures}.
\newblock \emph{Transactions of the Association for Computational Linguistics},
  8:743--758.

\bibitem[{Talmor et~al.(2019)Talmor, Herzig, Lourie, and
  Berant}]{talmor-etal-2019-commonsenseqa}
Alon Talmor, Jonathan Herzig, Nicholas Lourie, and Jonathan Berant. 2019.
\newblock \href {https://doi.org/10.18653/v1/N19-1421} {{C}ommonsense{QA}: A
  question answering challenge targeting commonsense knowledge}.
\newblock In \emph{Proceedings of the 2019 Conference of the North {A}merican
  Chapter of the Association for Computational Linguistics: Human Language
  Technologies, Volume 1 (Long and Short Papers)}, pages 4149--4158,
  Minneapolis, Minnesota. Association for Computational Linguistics.

\bibitem[{Talmor et~al.(2021)Talmor, Yoran, Catav, Lahav, Wang, Asai, Ilharco,
  Hajishirzi, and Berant}]{talmor2021multimodalqa}
Alon Talmor, Ori Yoran, Amnon Catav, Dan Lahav, Yizhong Wang, Akari Asai,
  Gabriel Ilharco, Hannaneh Hajishirzi, and Jonathan Berant. 2021.
\newblock \href {https://openreview.net/forum?id=ee6W5UgQLa} {Multimodal{QA}:
  complex question answering over text, tables and images}.
\newblock In \emph{International Conference on Learning Representations}.

\bibitem[{Thawani et~al.(2021)Thawani, Pujara, Ilievski, and
  Szekely}]{thawani-etal-2021-representing}
Avijit Thawani, Jay Pujara, Filip Ilievski, and Pedro Szekely. 2021.
\newblock \href {https://www.aclweb.org/anthology/2021.naacl-main.53}
  {Representing numbers in {NLP}: a survey and a vision}.
\newblock In \emph{Proceedings of the 2021 Conference of the North American
  Chapter of the Association for Computational Linguistics: Human Language
  Technologies}, pages 644--656, Online. Association for Computational
  Linguistics.

\bibitem[{Thorne et~al.(2021)Thorne, Yazdani, Saeidi, Silvestri, Riedel, and
  Halevy}]{DBLP:journals/corr/abs-2106-01074}
James Thorne, Majid Yazdani, Marzieh Saeidi, Fabrizio Silvestri, Sebastian
  Riedel, and Alon~Y. Halevy. 2021.
\newblock \href {http://arxiv.org/abs/2106.01074} {Database reasoning over
  text}.
\newblock \emph{CoRR}, abs/2106.01074.

\bibitem[{Wallace et~al.(2019)Wallace, Wang, Li, Singh, and
  Gardner}]{wallace-etal-2019-nlp}
Eric Wallace, Yizhong Wang, Sujian Li, Sameer Singh, and Matt Gardner. 2019.
\newblock \href {https://doi.org/10.18653/v1/D19-1534} {Do {NLP} models know
  numbers? probing numeracy in embeddings}.
\newblock In \emph{Proceedings of the 2019 Conference on Empirical Methods in
  Natural Language Processing and the 9th International Joint Conference on
  Natural Language Processing (EMNLP-IJCNLP)}, pages 5307--5315, Hong Kong,
  China. Association for Computational Linguistics.

\bibitem[{Wang et~al.(2020)Wang, Tsvetkov, and
  Neubig}]{wang-etal-2020-balancing}
Xinyi Wang, Yulia Tsvetkov, and Graham Neubig. 2020.
\newblock \href {https://doi.org/10.18653/v1/2020.acl-main.754} {Balancing
  training for multilingual neural machine translation}.
\newblock In \emph{Proceedings of the 58th Annual Meeting of the Association
  for Computational Linguistics}, pages 8526--8537, Online. Association for
  Computational Linguistics.

\bibitem[{Warstadt et~al.(2019)Warstadt, Cao, Grosu, Peng, Blix, Nie, Alsop,
  Bordia, Liu, Parrish, Wang, Phang, Mohananey, Htut, Jeretic, and
  Bowman}]{warstadt-etal-2019-investigating}
Alex Warstadt, Yu~Cao, Ioana Grosu, Wei Peng, Hagen Blix, Yining Nie, Anna
  Alsop, Shikha Bordia, Haokun Liu, Alicia Parrish, Sheng-Fu Wang, Jason Phang,
  Anhad Mohananey, Phu~Mon Htut, Paloma Jeretic, and Samuel~R. Bowman. 2019.
\newblock \href {https://doi.org/10.18653/v1/D19-1286} {Investigating
  {BERT}{'}s knowledge of language: Five analysis methods with {NPI}s}.
\newblock In \emph{Proceedings of the 2019 Conference on Empirical Methods in
  Natural Language Processing and the 9th International Joint Conference on
  Natural Language Processing (EMNLP-IJCNLP)}, pages 2877--2887, Hong Kong,
  China. Association for Computational Linguistics.

\bibitem[{Wolf et~al.(2020)Wolf, Debut, Sanh, Chaumond, Delangue, Moi, Cistac,
  Rault, Louf, Funtowicz, Davison, Shleifer, von Platen, Ma, Jernite, Plu, Xu,
  Scao, Gugger, Drame, Lhoest, and Rush}]{wolf2020huggingfaces}
Thomas Wolf, Lysandre Debut, Victor Sanh, Julien Chaumond, Clement Delangue,
  Anthony Moi, Pierric Cistac, Tim Rault, Rémi Louf, Morgan Funtowicz, Joe
  Davison, Sam Shleifer, Patrick von Platen, Clara Ma, Yacine Jernite, Julien
  Plu, Canwen Xu, Teven~Le Scao, Sylvain Gugger, Mariama Drame, Quentin Lhoest,
  and Alexander~M. Rush. 2020.
\newblock \href {http://arxiv.org/abs/1910.03771} {Huggingface's transformers:
  State-of-the-art natural language processing}.

\bibitem[{Xu et~al.(2019)Xu, Liu, Shen, Liu, and Gao}]{xu-etal-2019-multi}
Yichong Xu, Xiaodong Liu, Yelong Shen, Jingjing Liu, and Jianfeng Gao. 2019.
\newblock \href {https://doi.org/10.18653/v1/N19-1271} {Multi-task learning
  with sample re-weighting for machine reading comprehension}.
\newblock In \emph{Proceedings of the 2019 Conference of the North {A}merican
  Chapter of the Association for Computational Linguistics: Human Language
  Technologies, Volume 1 (Long and Short Papers)}, pages 2644--2655,
  Minneapolis, Minnesota. Association for Computational Linguistics.

\bibitem[{Yang et~al.(2021)Yang, Chen, Chen, and Cer}]{yang2021nt5}
Peng-Jian Yang, Ying~Ting Chen, Yuechan Chen, and Daniel Cer. 2021.
\newblock \href {http://arxiv.org/abs/2104.07307} {Nt5?! training t5 to perform
  numerical reasoning}.

\bibitem[{Yin et~al.(2020)Yin, Neubig, Yih, and Riedel}]{yin-etal-2020-tabert}
Pengcheng Yin, Graham Neubig, Wen-tau Yih, and Sebastian Riedel. 2020.
\newblock \href {https://doi.org/10.18653/v1/2020.acl-main.745} {{T}a{BERT}:
  Pretraining for joint understanding of textual and tabular data}.
\newblock In \emph{Proceedings of the 58th Annual Meeting of the Association
  for Computational Linguistics}, pages 8413--8426, Online. Association for
  Computational Linguistics.

\bibitem[{Yogatama et~al.(2019)Yogatama, de~Masson~d'Autume, Connor, Kocisky,
  Chrzanowski, Kong, Lazaridou, Ling, Yu, Dyer, and
  Blunsom}]{yogatama2019learning}
Dani Yogatama, Cyprien de~Masson~d'Autume, Jerome Connor, Tomas Kocisky, Mike
  Chrzanowski, Lingpeng Kong, Angeliki Lazaridou, Wang Ling, Lei Yu, Chris
  Dyer, and Phil Blunsom. 2019.
\newblock \href {http://arxiv.org/abs/1901.11373} {Learning and evaluating
  general linguistic intelligence}.

\bibitem[{Yu et~al.(2018)Yu, Dohan, Le, Luong, Zhao, and Chen}]{wei2018fast}
Adams~Wei Yu, David Dohan, Quoc Le, Thang Luong, Rui Zhao, and Kai Chen. 2018.
\newblock \href {https://openreview.net/forum?id=B14TlG-RW} {Fast and accurate
  reading comprehension by combining self-attention and convolution}.
\newblock In \emph{International Conference on Learning Representations}.

\bibitem[{Yu et~al.(2021)Yu, Wu, Lin, bailin wang, Tan, Yang, Radev, richard
  socher, and Xiong}]{yu2021grappa}
Tao Yu, Chien-Sheng Wu, Xi~Victoria Lin, bailin wang, Yi~Chern Tan, Xinyi Yang,
  Dragomir Radev, richard socher, and Caiming Xiong. 2021.
\newblock \href {https://openreview.net/forum?id=kyaIeYj4zZ} {Gra{PP}a:
  Grammar-augmented pre-training for table semantic parsing}.
\newblock In \emph{International Conference on Learning Representations}.

\bibitem[{Zhang et~al.(2020)Zhang, Zhang, Zhang, and
  Søgaard}]{zhang2020worstcaseaware}
Sheng Zhang, Xin Zhang, Weiming Zhang, and Anders Søgaard. 2020.
\newblock \href {http://arxiv.org/abs/2009.11138} {Worst-case-aware curriculum
  learning for zero and few shot transfer}.

\bibitem[{Zhao et~al.(2019)Zhao, Zhu, and Yu}]{zhao-etal-2019-data}
Zijian Zhao, Su~Zhu, and Kai Yu. 2019.
\newblock \href {https://doi.org/10.18653/v1/D19-1375} {Data augmentation with
  atomic templates for spoken language understanding}.
\newblock In \emph{Proceedings of the 2019 Conference on Empirical Methods in
  Natural Language Processing and the 9th International Joint Conference on
  Natural Language Processing (EMNLP-IJCNLP)}, pages 3637--3643, Hong Kong,
  China. Association for Computational Linguistics.

\end{thebibliography}
\bibliographystyle{acl_natbib}

\appendix

\section{Supplemental Material}
\label{sec:appendix}


\subsection{Data Generation}\label{sec:sup-dg}

\begin{figure}[ht]
  \includegraphics[width=1.0\columnwidth]{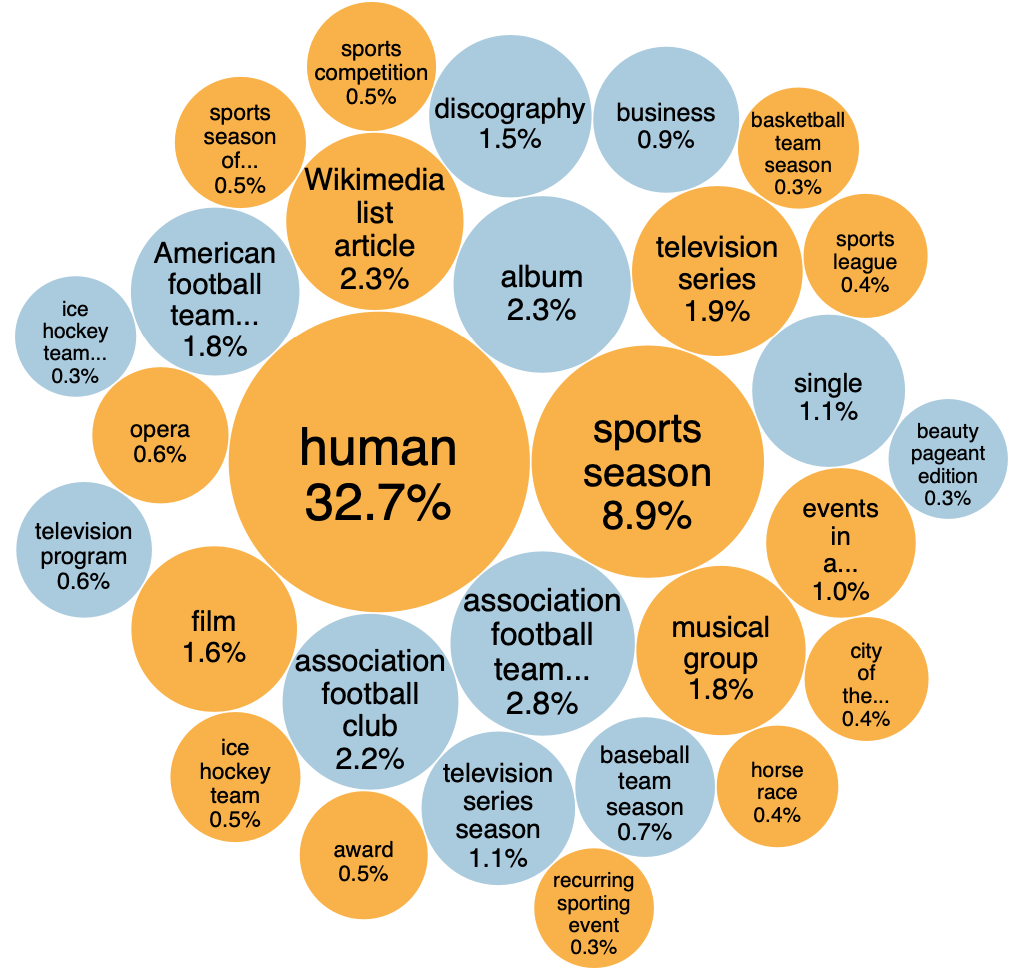}
  \caption{The most frequent categories of our Wikipedia pages and their frequency.
  }
  ~\label{fig:domain-diversity}
\end{figure}

Table~\ref{tab:syn-num-examples} contains the number of generated examples for every EG. During data generation, we randomly generate at most $10$ examples for each EG and table. Table~\ref{tab:examples-full-context} contains examples for generated $(q,c,a)$ triplets, including the full context $c$. Fig.~\ref{fig:domain-diversity} presents the most common categories of the Wikipedia pages from which we scraped our tables.

\begin{table}[ht!]
    \centering
        \tiny
        \begin{tabular}{lc} \toprule
        \emph{EG} & \emph{\# Questions} \\ \midrule
        2-Hop composition & 277,069 \\ 
        3-Hop composition & 364,772 \\ 
        Conjunction & 353,738\\ 
        Only quantifier & 522,071\\
        Most quantifier & 94,180 \\ 
        Every quantifier & 16,693 \\ 
        Number comparison & 410,749 \\
        Temporal comparison & 453,499 \\
        Number boolean comparison & 410,749 \\
        Temporal boolean comparison & 470,694 \\
        Number superlatives & 125,144 \\
        Temporal superlatives & 80,884 \\
        Arithmetic superlatives & 183,892 \\
        Arithmetic addition & 86,969 \\
        Counting & 484,471 \\
        Date difference & 452,061  \\ \midrule
        Total & 4,787,635 \\ \bottomrule
        
        \end{tabular}
        \caption{Number of examples generated by each EG.
        }
        \label{tab:syn-num-examples}
\end{table}

\begin{table*}
    \begin{center}
        \tiny
        \begin{tabular}{l|l|l|l} \toprule
        {\emph{EG}} &  {\emph{Question}} & {\emph{Context}} & {\emph{Answer}}  \\ \midrule
        
        3-hop & What was the Result(s) when the  & In League Cup of 1990-91 Chelsea F.C. season: The attendance when the round was R2 1st Leg was 5,666.  & 2-1 \\
         Composition & Round was R4 in League Cup of  & The result when the date was 6 November 1990 was 3-2. The date when the attendance was 34,669 was 27 & \\
         & 1990-91 Chelsea F.C. season? & February 1991. The attendance when the round was QF was 34,178. The date when the attendance was  & \\
         && 34,074 was 24 February 1991. The date when the attendance was 16,085 was 6 November 1990. The & \\
         && attendance when the round was R3 was 16,699. \textbf{The date when the attendance was 9,789 was 28 November} & \\
         &&\textbf{1990.} \textbf{The result when the date was 28 November 1990 was 2-1.} The result when the date was 31 October &\\ && 1990 was 0-0. The attendance when the round was QFR was 33,861. The result when the date was 16 &\\
         && January 1991 was 0-0. \textbf{The attendance when the round was R4 was 9,789.} The result when the date was &\\
         &&10 October 1990 was 4-1 (won 9-1 on agg). The date when the attendance was 5,666 was 26 September 1990.\\ \midrule
         
        Numerical & Which opponent has the highest & In League Cup of 1990-91 Chelsea F.C. season: The attendances when the opponent was Tottenham Hotspur & Sheffield \\
        Superlatives & attendance in League Cup of &  were 34,178 and 33,861. The attendances when the opponent was Sheffield Wednesday were 34,669 and & Wednesday \\
        & 1990-91 Chelsea F.C. season? & 34,074. The attendance when the opponent was Oxford United was 9,789. The attendances when the & \\
        && opponent was Portsmouth were 16,699 and 16,085. The attendances when the opponent was Walsall were &\\
        && 5,666 and 10,037. &\\
        \bottomrule
        
        \end{tabular} 
        \caption{
        Examples for generated $(q,c,a)$ triplets. The examples were generated from the table shown in fig.~\ref{fig:intro}. The gold facts for the composition question are indicated. The numerical superlatives question requires reasoning over all the facts in the context.
        }
        \label{tab:examples-full-context}
    \end{center}
\end{table*}

\subsection{Training}\label{sec:sup-training}
\paragraph{Error sampling}
Alg.~\ref{alg:error} provides our error sampling algorithm.

\begin{algorithm}
    \small
	\caption{Error Sampling($t$)} 
	\textbf{Input:} training time $t$
	\begin{algorithmic}[1]
         
	    \For {$s \in \mathcal{S}$}
        
    		    \State $P_\textit{tasks}[s]$ $\leftarrow{}$ $1.0-\textit{Acc}_s(t)$

		\EndFor
	\State $P_\textit{tasks} \leftarrow{} P_\textit{tasks}/\| P_\textit{tasks} \|_{1}$ 
	\end{algorithmic} 
	\label{alg:error}
\end{algorithm}

\paragraph{Momentum Sampling}

For momentum sampling we use a window size of $w=4$, a smoothing factor of $k=2$, and sample at least $\epsilon=0.002$ examples from every task when training \rem{-Base} and \rem{-Large}.

\subsection{Experimental Setup}\label{sec:sup-setup}

\paragraph{Original pre-training task}
In order to avoid catastrophic forgetting \cite{DBLP:journals/corr/KirkpatrickPRVD16}, we continue training with the span-corruption objective introduced in \cite{2020t5}, over sequences of length 256 from the English Wikipedia.

\paragraph{Implementation details}
We fine-tune all our experiments on one RTX8000 (48GB) or RTX3090 (24GB) GPU.
We use the \tfive{} model from \url{https://huggingface.co/transformers/model_doc/t5.html} \cite{wolf2020huggingfaces}.

\begin{table}[H]\resizebox{1.0\columnwidth}{!}{
\centering
        \tiny
        \begin{tabular}{ll|ccc|c} \toprule
        \emph{Experiment}  & \emph{Size} & \emph{LR} & \emph{Batch Size} &  \emph{GAS} & \emph{Epochs}\\ \midrule 
        \rem & Base & 1e-4 & 64 & 1 & 50 \\
        \rem & Large & 1e-4 & 18 & 4 & 36 \\ \midrule 
        DROP & Base & 1e-4 & 20 & 1 & 20 \\
        IIRC & Base & 1e-4 & 20 & 1 & 60 \\
        IIRC$_\textrm{oracle}$ & Base & 1e-4 & 20 & 1 & 60 \\
        MMQA & Base &  1e-4 & 6 & 3 & 20  \\
        DROP & Large & 5e-5 & 16 & 2 & 20 \\
        IIRC & Large & 5e-5 & 16 & 2 & 60 \\
         IIRC$_\textrm{oracle}$ & Large & 5e-5 & 16 & 2 & 60 \\
        MMQA & Large & 1e-4 & 2 & 16 & 10\\

        \bottomrule 
         
        \end{tabular}} 
\caption{Hyper-parameters used in all experiments, LR and GAS refer to learning-rate and gradient accumulation steps. In our \rem{} experiments, epochs refer to the number of steps between evaluations, which is set to $5,000$ and $2,500$ for our base and large experiments, which leads to $250,000$ and $90,000$ optimization steps, respectively.} 
\label{tab:hyperparams}
\end{table}

\subsection{Experimental Results}\label{sec:sup-results}

\begin{figure*}[ht!]
  \centering
  \includegraphics[width=2.0\columnwidth]{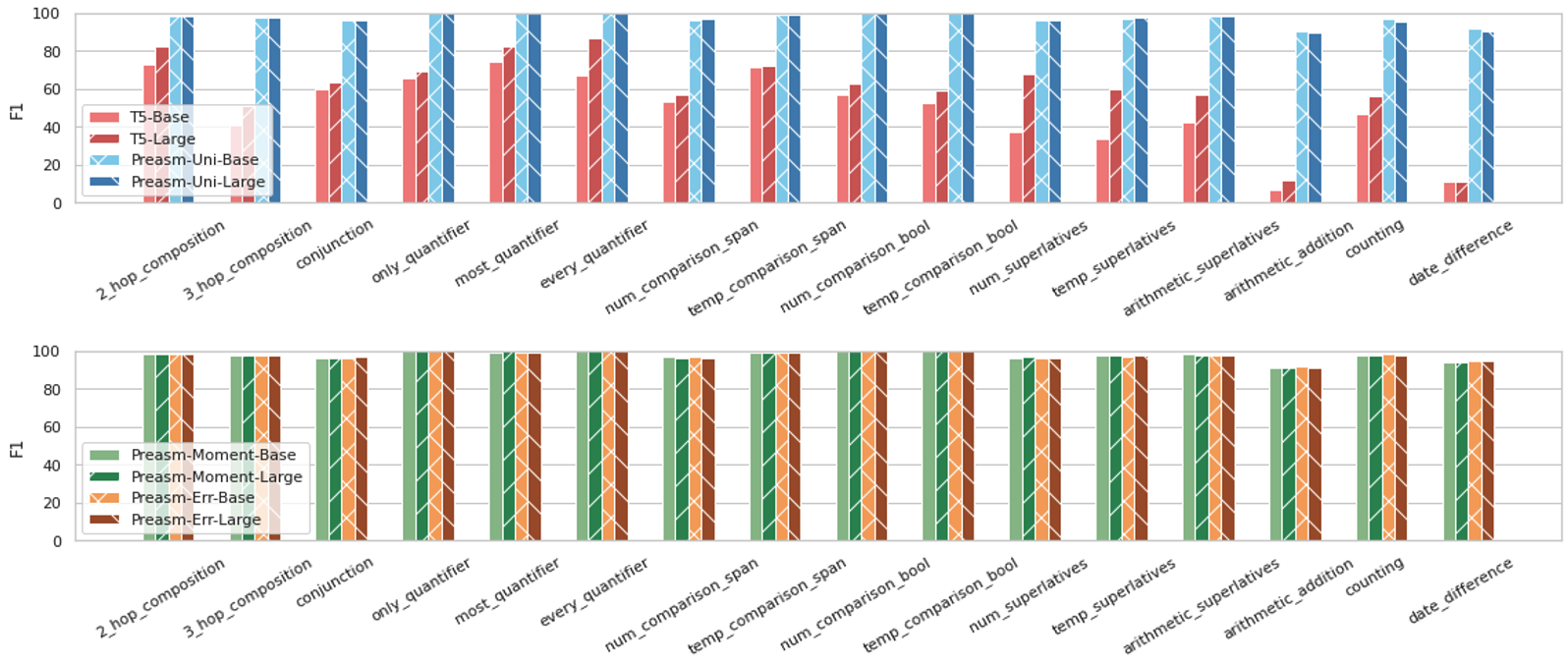}
  \caption{
    \fone{} for every task, for \tfive{} and \rem{}. The results for \tfive{} were received by training in a few shot manner on 32 examples for 200 steps, as suggested in \cite{ram2021fewshotqa}.
  }
  ~\label{fig:syn-t5-vs-rem}
\end{figure*}

Fig.~\ref{fig:syn-t5-vs-rem} shows the results for \tfive{} and \rem{} on $D_\textit{syn}$ for both model sizes. \tfive{-Large} outperforms \tfive{-Base} on most tasks, suggesting that skills such as comparison and superlatives may have been picked up better during pre-training. However on tasks such as date difference and arithmetic addition the results \tfive{-Large} are very low, at around $10$ \fone{}. Our \rem{} models significantly outperforms \tfive{} on all tasks.

\end{document}